\documentclass{article}
\usepackage{graphicx}
\PassOptionsToPackage{numbers, compress}{natbib}
\usepackage{multirow}
\usepackage{wrapfig}   
\usepackage{caption}
\usepackage{float}
\usepackage{tikz}
 \usepackage[preprint]{neurips_2026}

\usepackage[table]{xcolor}
\usepackage[utf8]{inputenc} 
\usepackage[T1]{fontenc}    
\usepackage{hyperref}       
\usepackage{url}            
\usepackage{booktabs}       
\usepackage{amsfonts}       
\usepackage{nicefrac}       
\usepackage{microtype}      
\usepackage{xcolor}         
\usepackage{amsmath}        
\usepackage{amssymb}        
\usepackage{algorithm}      
\usepackage{algpseudocode}  

\title{DriveFuture: Future-Aware Latent World Models for Autonomous Driving}

\author{
\textbf{Yufeng Hong}\textsuperscript{\rm 1$\dagger$},
\textbf{Xiaotian Zhou}\textsuperscript{\rm 1$\dagger$},
\textbf{Yingyan Li}\textsuperscript{\rm 2},
\textbf{Xiangpo Zhou}\textsuperscript{\rm 3},
\textbf{Lin Liu}\textsuperscript{\rm 4},\\
\textbf{Yadan Luo}\textsuperscript{\rm 5},
\textbf{Shaoqing Xu}\textsuperscript{\rm 6},
\textbf{Lei Yang}\textsuperscript{\rm 7},
\textbf{Ziying Song}\textsuperscript{\rm 8,7$\ast$}
\\
\\
\textsuperscript{\rm 1}Beijing Institute of Technology \textsuperscript{\rm 2}Institute of Automation, Chinese Academy of Sciences\\
\textsuperscript{\rm 3}Beihang University
\textsuperscript{\rm 4}Beijing Jiaotong University
\textsuperscript{\rm 5}The University of Queensland\\
\textsuperscript{\rm 6}University of Macau
\textsuperscript{\rm 7}Nanyang Technological University \\
\textsuperscript{\rm 8}School of Artificial Intelligence ( School of Software), Yanshan University\\
\\
\textsuperscript{\rm $\dagger$}Equal contribution.
\textsuperscript{\rm $\ast$}Corresponding author.
}

\begin{document}

\maketitle


\begin{abstract}
Existing latent world models for autonomous driving have opened a promising path toward future-aware driving intelligence. However, they typically treat future latent states as prediction targets or auxiliary signals, rather than directly conditioning trajectory planning. This can entangle current and future features in latent space.
In this work, we propose DriveFuture, a future-aware latent world modeling framework for autonomous driving that explicitly learns planning-oriented foresight by conditioning the current latent state modeling process on future world states. Specifically, during training, the model first predicts future latent world states from the current latent state and ego action, and then refines the prediction against the ground-truth future latent state via cross-attention. The resulting future-aware latent serves as an explicit condition for a diffusion-based trajectory planner. During inference, DriveFuture conditions on the predicted future latent state instead of the ground-truth future state. DriveFuture achieves SOTA performance on the public NAVSIM benchmarks~\cite{navsim}, reaching \textbf{55.5} EPDMS on  NAVSIM-v2 {\textcolor{blue}{\textit{navhard}}}, \textbf{89.9} EPDMS on  NAVSIM-v2 {\textcolor{blue}{\textit{navtest}}}, and \textbf{90.7} PDMS on  NAVSIM-v1 {\textcolor{blue}{\textit{navtest}}}, respectively. These results suggest that the key to latent world modeling lies not merely in simulating future states, but more importantly in conditioning current decision-making on future states. Notably, as of April 2026, DriveFuture ranks \textbf{1st} on the 
\href{https://huggingface.co/spaces/AGC2025/e2e-driving-navhard}{NAVSIM-v2 {\textcolor{blue}{\textit{navhard}}}} 
leaderboard and achieves SOTA performance on 
\href{https://huggingface.co/spaces/AGC2024-P/e2e-driving-navtest}{NAVSIM-v1 {\textcolor{blue}{\textit{navtest}}}}.
\end{abstract}

\section{Introduction}









In recent years, world models  have gradually emerged as a central direction in autonomous driving and are widely regarded as one of the most promising technical paradigms for achieving higher-level driving intelligence \cite{jia2025progressivesurvey,feng2025survey,tu2025role,song2024robustness,luo2026unleashing}. Unlike conventional approaches \cite{uniad,liao2025diffusiondrive,sun2025sparsedrive,xing2025goalflow,liu2025guideflow,sun2025focalad,momad,suna2025minddrive,chen2024vadv2,li2025recogdrive}, which directly map the current state to driving actions, world models \cite{jia2025progressivesurvey,li2024enhancinglaw} learn the latent dynamics of the driving environment. This allows the system to anticipate future outcomes under different actions and make more informed planning decisions. By reasoning about how the world may evolve before selecting an action, this paradigm shifts autonomous driving from reactive control toward future-aware decision-making. As a result, world models are evolving from an auxiliary modeling tool into a key foundation for scalable and generalizable autonomous driving.

Among existing world-modeling paradigms, latent world models are particularly promising for autonomous driving for three reasons. First, they learn dynamics in compact latent spaces, avoiding costly pixel-level generation. Second, latent representations can abstract away low-level visual details and focus on planning-relevant scene structure, interactions, and action consequences. Third, their temporally structured representations are naturally suited for long-horizon planning. This line of work has rapidly evolved from latent future prediction \cite{li2024enhancinglaw} to intention-aware latent planning \cite{zheng2025world4drive}, planning-oriented representation refinement with reinforcement fine-tuning \cite{yang2026worldrft}, and more recent unifications with VLA/planners and policy scaling \cite{liu2026driveworld,li2025drivevla,xia2025drivelaw,min2024driveworld}. These developments indicate that latent world models are no longer merely an efficient substitute for observation-space simulation, but are increasingly becoming a general representation substrate for scalable autonomous driving.

\begin{figure}[t]
    \centering
    \includegraphics[width=1\linewidth]{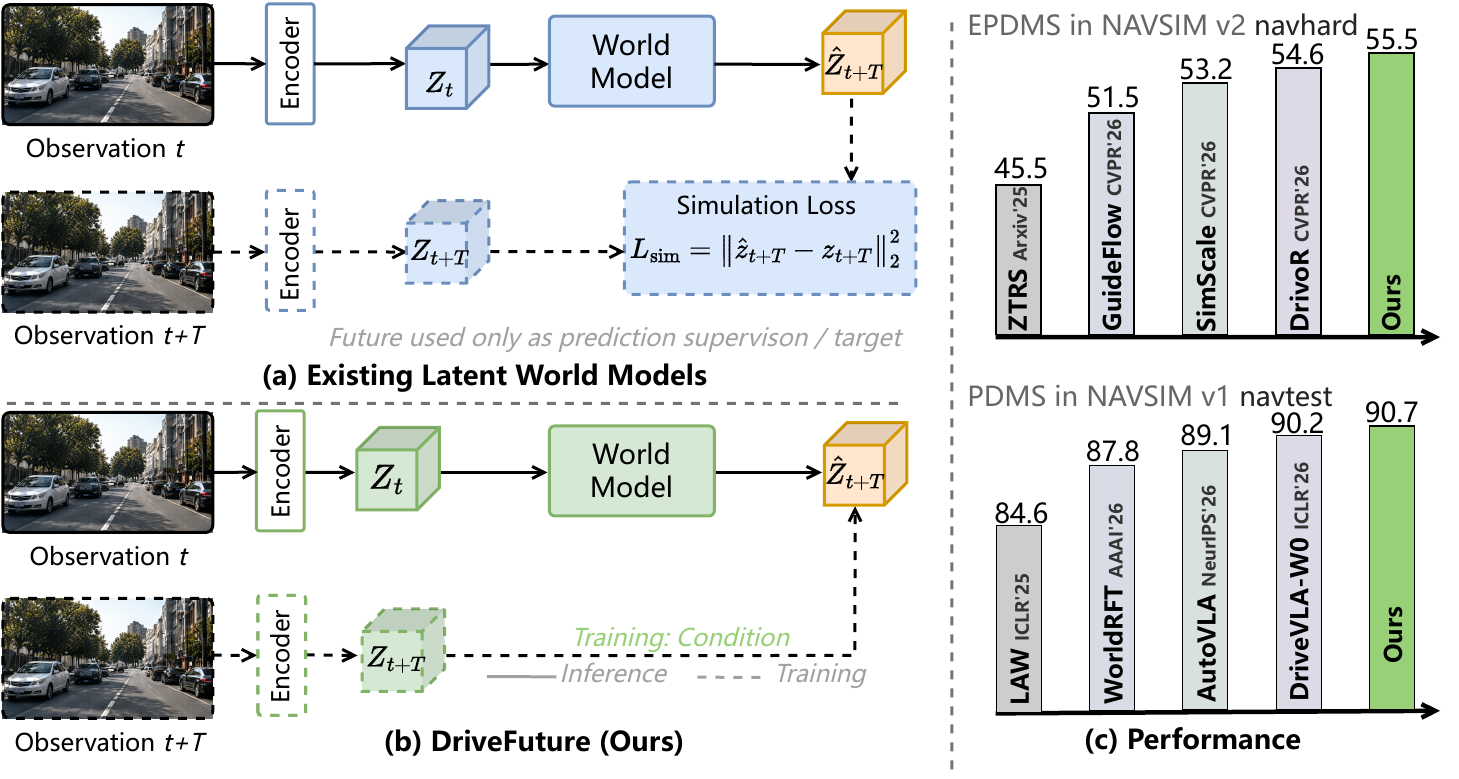}
    \caption{
Motivation of \textbf{DriveFuture}. 
(a) Existing latent world models \cite{li2024enhancinglaw,zheng2025world4drive,yang2026worldrft,min2024driveworld,liu2026driveworld,li2025drivevla,xia2025drivelaw}  primarily simulate future latent states and use them as prediction targets or supervision signals, without explicitly shaping the current representation for planning. 
(b) DriveFuture uses future latent states as direct conditions for the planning process. It adopts GT future states during training and predicted future states during inference, enabling future-aware trajectory planning. 
(c) DriveFuture achieves SOTA performance on the public leaderboards of 
\href{https://huggingface.co/spaces/AGC2025/e2e-driving-navhard}{NAVSIM-v2 {\textcolor{blue}{\textit{navhard}}}} 
and 
\href{https://huggingface.co/spaces/AGC2024-P/e2e-driving-navtest}{NAVSIM-v1 {\textcolor{blue}{\textit{navtest}}}}.
}
    \label{fig:motivation}
    \vspace{-0.3cm}
\end{figure}

Although latent world models \cite{li2024enhancinglaw,zheng2025world4drive,yang2026worldrft,min2024driveworld,liu2026driveworld,li2025drivevla,xia2025drivelaw} provide an efficient paradigm for future modeling in autonomous driving, existing methods primarily focus on \emph{future state prediction} rather than \emph{current decision representation}. Specifically, as shown in Fig. \ref{fig:motivation}, they typically treat future latent states as prediction targets or auxiliary signals, rather than using them as explicit conditions to shape the current decision-making process, which leads to entanglement between current and future features in latent space. As a result, these models are better at \emph{simulating the future} than at learning planning-oriented decision representations, thereby limiting the ability of future information to guide current decision-making.

We argue that the role of future information in autonomous driving is not merely to predict how future scenes will evolve, but more importantly to endow the system with a decision-making perspective akin to that in \emph{The Terminator}, namely, \emph{bringing future knowledge back to the present}: using awareness of future consequences at the current moment to reshape the understanding of the present scene and the resulting decision. An ideal autonomous driving system, therefore, should not be merely a future predictor, but should instead allow future states to act as conditions that influence the current decision-making process in reverse, thereby enabling more foresighted planning. Inspired by this observation, we propose \textbf{DriveFuture}, a future-aware latent world modeling framework for autonomous driving. By conditioning the current decision-making process on future world states, DriveFuture explicitly learns planning-oriented \emph{foresight}. Unlike prior approaches that treat future latent states solely as prediction targets, DriveFuture regards them as structured conditions for shaping the current decision-making process.

Our framework follows a simple future-aware latent world modeling pipeline. During training, DriveFuture first predicts future latent world states from the current latent state and ego action, and then uses the ground-truth (GT) future latent world state as a grounding signal for the predicted future latent, which then conditions the diffusion-based trajectory planner. This process enables the planner to leverage future semantics for more informed trajectory generation. During inference, GT future latent states are no longer available, and the model instead uses predicted future latent states as conditions, enabling closed-loop planning without future annotations. DriveFuture establishes a unified future-aware latent dynamics mechanism for both training and inference. We evaluate DriveFuture on the public \textit{NAVSIM} benchmarks~\cite{navsim} and observe substantial performance gains. Specifically, DriveFuture achieves \textbf{55.5}, \textbf{89.9}, and \textbf{90.7} EPDMS on \textit{ NAVSIM-v2 navhard}, \textit{ NAVSIM-v2 navtest}, and \textit{ NAVSIM-v1 navtest}, respectively. These results show that future-aware latent world modeling can significantly improve planning quality in challenging closed-loop driving scenarios.
In summary, the main contributions of this work are as follows:

\begin{itemize}
    \item We identify a key limitation of existing latent world models for autonomous driving: they mainly focus on future-state simulation, while failing to fully exploit future states as conditions for current decision-making, which leads to current--future feature entanglement.
    \item We propose \textbf{DriveFuture}, a future-aware latent world modeling framework that conditions the current decision-making process on future world states to learn planning-oriented \emph{foresight}. It adopts a unified training--inference paradigm, using GT future latent states as conditions during training and predicted future latent states during inference.
    \item We achieve SOTA results on the public NAVSIM benchmarks~\cite{navsim}, demonstrating the effectiveness of future-conditioned latent world modeling. Notably, as of April 2026, DriveFuture ranks \textbf{1st} on the 
\href{https://huggingface.co/spaces/AGC2025/e2e-driving-navhard}{NAVSIM-v2 {\textcolor{blue}{\textit{navhard}}}} leaderboard with \textbf{55.5} EPDMS, and achieves SOTA performance on 
\href{https://huggingface.co/spaces/AGC2024-P/e2e-driving-navtest}{NAVSIM-v1 {\textcolor{blue}{\textit{navtest}}}} with \textbf{90.7} PDMS.

\end{itemize}

\section{Related Work}

\subsection{World Models for Autonomous Driving}
Recent research on world modeling for autonomous driving can be broadly grouped into several directions. The first line of work explicitly models future scene evolution in the observation space, including multi-view driving video generation and world simulation methods \cite{hu2023gaia1,wang2024drivedreamer,gao2024vista,zhang2025epona,yang2025resim,yang2026consisdrive,wang2024driving,jia2023adriver,wang2025adawm}, which emphasize high-fidelity future rollout, controllable scene evolution, long-horizon generation, and reliable simulation for downstream planning. The second line models the evolution of 3D/4D world states in occupancy space \cite{zheng2024occworld,yang2025driving,wang2024occsora,wei2024occllama}, which characterize future worlds from the perspectives of 3D occupancy world modeling, vision-centric 4D occupancy forecasting, 4D occupancy generation, and unified occupancy-language-action modeling, respectively. The third line performs world modeling in more compact BEV or structured state spaces \cite{zhang2024bevworld,wote}, which support future evaluation and trajectory selection through a unified BEV latent space or a BEV world model. More recently, world models have also been increasingly integrated with planner- or VLA-oriented frameworks, as exemplified by LAW \cite{li2024enhancinglaw}, World4Drive \cite{zheng2025world4drive}, WorldRFT \cite{yang2026worldrft}, DriveWorld-VLA \cite{liu2026driveworld}, DriveVLA-W0 \cite{li2025drivevla}, and DriveLaW \cite{xia2025drivelaw}, suggesting a clear trend toward tighter coupling between world modeling, trajectory planning, and scalable autonomous driving systems. Although these methods have demonstrated strong potential for future scene modeling and planning support, they typically involve high modeling complexity and are prone to error accumulation in observation-space or dense-space reconstruction.

\subsection{End-to-End Autonomous Driving.}
End-to-end autonomous driving (E2E-AD) methods map raw sensor observations directly to vehicle controls or planned trajectories. Early E2E-AD methods  \cite{chitta2022transfuser,uniad,jiang2023vad,chen2024vadv2,xu2024m2da,guo2024uad,sun2025sparsedrive,li2024hydra,momad,sun2025focalad,liu2025fump,zheng2024genad,li2025recogdrive,zhang2025bridging,liu2026reinforced}, mainly focus on improving scene representation, multi-modal fusion, and planning stability within unified end-to-end frameworks. More recent approaches increasingly introduce generative planning mechanisms to better capture multi-modal futures and long-horizon behaviors. Representative examples include DiffusionDrive \cite{liao2025diffusiondrive}, DIVER \cite{song2025diver}, GoalFlow \cite{xing2025goalflow}, GuideFlow \cite{liu2025guideflow}, and GTRS \cite{GTRS}, which improve trajectory diversity, scoring, controllability, or reasoning ability through diffusion, flow matching, reinforcement learning, or stronger trajectory evaluation. These advances suggest that strong E2E-AD performance increasingly depends on reasoning over possible future evolutions rather than solely reacting to the current scene. However, in most existing systems, future information is introduced only at the output level, e.g., by generating multiple candidate trajectories or scoring future outcomes after the current representation has already been formed. As a result, they improve trajectory generation or selection, but do not fundamentally reshape how the current latent state itself is learned for planning.
Instead, DriveFuture leverages future latent states as direct conditions for trajectory planning, explicitly aligning the decision-making process with downstream objectives.



\begin{figure}[t]
    \centering
    \includegraphics[width=1\linewidth]{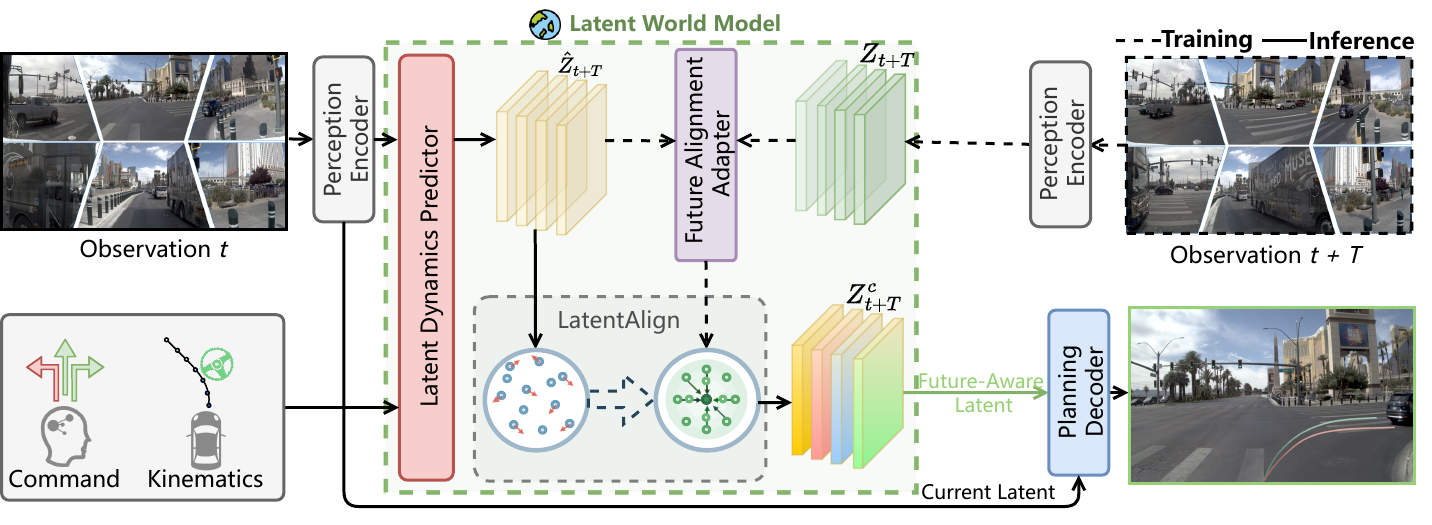}
    \caption{
\textbf{Overview of DriveFuture.}
Multi-view observations at time $t$ are encoded by a shared Perception Encoder into a scene latent $\mathbf{Z}_t$.
The \textit{Latent Dynamics Predictor} conditions on $\mathbf{Z}_t$ and a tokenised trajectory intent to produce a predicted future latent $\hat{\mathbf{Z}}_{t+T}$.
During training, the future observation at $t{+}T$ is encoded by the same Perception Encoder into $\mathbf{Z}_{t+T}$, the \textit{Future Alignment Adapter} grounds $\hat{\mathbf{Z}}_{t+T}$ via cross-attention against $\mathbf{Z}_{t+T}$, yielding the future-aware latent $\mathbf{Z}^c_{t+T}$.
\textit{LatentAlign} anneals the planning condition from $\mathbf{Z}^c_{t+T}$ towards $\hat{\mathbf{Z}}_{t+T}$ over training, closing the train--inference gap.
At inference, the adapter is bypassed, the \textit{Planning Decoder} consumes $\mathbf{Z}_t$ and $\hat{\mathbf{Z}}_{t+T}$ as dual conditioning contexts to iteratively denoise the ego trajectory.
}
    \label{fig:method}
    \vspace{-0.3cm}
\end{figure}





\section{Method}
\label{sec:method}
\noindent
The central principle of DriveFuture is to leverage predicted future latents to condition and enhance trajectory planning. As shown in Fig.~\ref{fig:method}, our framework is built upon three core modules: the Latent Dynamics Predictor~(Sec.~\ref{sec:method:wm}), the Future Alignment Adapter~(Sec.~\ref{sec:method:align}), and the Planning Decoder~(Sec.~\ref{sec:method:dec}). We further detail the Training~(Sec.~\ref{sec:method:train}) and Inference~(Sec.~\ref{sec:method:infer}) pipelines.

\subsection{Latent Dynamics Predictor}
\label{sec:method:wm}

To provide the planning decoder with a compact, action-controllable future signal while avoiding the computational overhead of observation-space prediction, we instantiate a trajectory-conditioned latent predictor in a shared BEV latent space.

\paragraph{Trajectory-Conditioned Latent Prediction.}
Let $\phi_{\mathrm{enc}}$ denote the shared BEV encoder, which maps the multi-view observation $\mathbf{I}_{t}$ and ego status $\mathbf{s}_{t}$ into a scene latent $\mathbf{Z}_{t}$ comprising $N$ BEV tokens and one ego status token. Let $\phi_{\tau}$ encode an absolute trajectory $\boldsymbol{\tau}\!=\!\{(x_{k},y_{k},\theta_{k})\}_{k=1}^{T}$ into a token sequence $\mathbf{E}_{\tau}\!=\!\phi_{\tau}(\boldsymbol{\tau})$ via a normalised differential representation $(\Delta x,\Delta y,\sin\theta,\cos\theta)$, a linear projection, and a temporal positional embedding. A learnable future query bank $\mathbf{Q}_{f}\!\in\!\mathbb{R}^{K\times d}$ attends to the joint context through a stack of transformer decoder layers $f_{\psi}$, producing
\begin{equation}
\hat{\mathbf{Z}}_{t+T}
\;=\;
f_{\psi}\!\left(\mathbf{Q}_{f}\,;\;[\mathbf{Z}_{t}\,\Vert\,\mathbf{E}_{\tau}]\right).
\label{eq:wm_forward}
\end{equation}
We refer to $\hat{\mathbf{Z}}_{t+T}$ as the \emph{trajectory-conditioned latent prediction}: it describes the scene latent at horizon $T$ under the hypothesis that the ego enacts $\boldsymbol{\tau}$, and serves as the substrate on which the Future Alignment Adapter and the Planning Decoder operate.


\paragraph{Conditioning Source Randomisation.}
The predictor in Eq.~\eqref{eq:wm_forward} requires a trajectory $\boldsymbol{\tau}$ that is unavailable at inference time, and classifier-free guidance (CFG) additionally demands calibrated outputs under absent or coarse trajectory inputs. Both requirements are addressed jointly by drawing the trajectory token sequence from three sources during training, with probabilities $\{p_{\mathrm{gt}}, p_{\mathrm{kin}}, p_{\varnothing}\}$:
\begin{equation}
\mathbf{E}_{\tau}\;\sim\;
\begin{cases}
\phi_{\tau}(\boldsymbol{\tau}^{\mathrm{gt}}), & \text{w.p.}\;p_{\mathrm{gt}},\\[2pt]
\phi_{\tau}(\boldsymbol{\tau}^{\mathrm{kin}}), & \text{w.p.}\;p_{\mathrm{kin}},\\[2pt]
\mathbf{E}^{\varnothing}, & \text{w.p.}\;p_{\varnothing},
\end{cases}
\label{eq:multi_src_dropout}
\end{equation}
where $\mathbf{E}^{\varnothing}$ is a learned null token sequence and $\boldsymbol{\tau}^{\mathrm{kin}}$ is a coarse but stable kinematic surrogate produced by constant-acceleration extrapolation from the ego velocity $(v_{x},v_{y})$ and acceleration $(a_{x},a_{y})$:
\begin{equation}
x_{k}\!=\!v_{x}t_{k}\!+\!\tfrac{1}{2}a_{x}t_{k}^{2},\quad
y_{k}\!=\!v_{y}t_{k}\!+\!\tfrac{1}{2}a_{y}t_{k}^{2},\quad
\theta_{k}\!=\!\operatorname{atan2}\!\left(v_{y}\!+\!a_{y}t_{k},\,v_{x}\!+\!a_{x}t_{k}\right),
\label{eq:kin_extrap}
\end{equation}
with $t_{k}\!=\!k\Delta t$. The three branches expose the predictor to fine, coarse, and unconditional regimes during training, so that the Tweedie-based and kinematic-based guidance terms used in Sec.~\ref{sec:method:infer} share a well-trained baseline.

\subsection{Future Alignment Adapter}
\label{sec:method:align}


The trajectory diffusion loss reaches the latent dynamics predictor only through the planning decoder, which is a weak and indirect signal under which $\hat{\mathbf{Z}}_{t+T}$ is free to drift from the true future semantics in the early stages of training. The Future Alignment Adapter introduces a grounded but lightweight signal that anchors the predicted future latent to the actual future scene, while avoiding dense reconstruction targets that would tie the latent to appearance details unrelated to planning.

\paragraph{Cross-Attention Grounding.}
During training, the future-step observation $\mathbf{I}_{t+T}$ is passed through the same encoder $\phi_{\mathrm{enc}}$ used for the current frame, with gradients stopped to prevent the future branch from acting as a shortcut, yielding $\mathbf{Z}_{t+T}\!=\!\mathrm{sg}(\phi_{\mathrm{enc}}(\mathbf{I}_{t+T}))$. We use $\mathbf{Z}_{t+T}$ as the empirical proxy for the GT future latent state. The adapter is a single multi-head cross-attention layer with $\hat{\mathbf{Z}}_{t+T}$ as queries and $\mathbf{Z}_{t+T}$ as keys/values:
\begin{equation}
\tilde{\mathbf{Z}}_{t+T}
\;=\;
\mathrm{MHA}\!\left(\,
\mathrm{LN}(\hat{\mathbf{Z}}_{t+T}),\;\;
\mathbf{Z}_{t+T},\;\;
\mathbf{Z}_{t+T}
\,\right).
\label{eq:latentalign}
\end{equation}
The query side preserves the $K$-token interface used by the Planning Decoder, so the adapter never enlarges the conditioning footprint regardless of the spatial resolution of the future BEV. Because the future observation is unavailable for a subset of the training samples (e.g., at sequence boundaries), let $\mathcal{S}_{+}$ denote the indices that admit a future frame, the planning condition is
\begin{equation}
\mathbf{Z}_{t+T}^{c,(i)}\;=\;
\begin{cases}
\tilde{\mathbf{Z}}_{t+T}^{(i)},& i\in\mathcal{S}_{+},\\[2pt]
\hat{\mathbf{Z}}_{t+T}^{(i)},& i\notin\mathcal{S}_{+},
\end{cases}
\label{eq:cfut_mask}
\end{equation}

which falls back to the unrefined forecast on samples without future observations. At inference time, $\mathbf{Z}_{t+T}$ is unavailable and the adapter is bypassed, so $\mathbf{Z}_{t+T}^{c}=\hat{\mathbf{Z}}_{t+T}$.

\paragraph{LatentAlign.}
To bridge the train--inference gap caused by the unavailability of the grounded future latent $\mathbf{Z}_{t+T}^{c}$ at inference time, we propose LatentAlign, a sigmoid-based annealing schedule that smoothly transitions the planning condition from the grounded future latent to the self-predicted forecast over the course of training:
\begin{equation}
\alpha(e)
\;=\;
1\;-\;\sigma\!\left(\beta\,(e - e_{0})\right),
\qquad
\tilde{\mathbf{Z}}_{t+T}^{c}
\;=\;
\alpha(e)\,\mathbf{Z}_{t+T}^{c}\;+\;\bigl(1-\alpha(e)\bigr)\,\hat{\mathbf{Z}}_{t+T},
\label{eq:latentalign_anneal}
\end{equation}
where $\sigma(\cdot)$ is the logistic function. Early in training ($\alpha\!\to\!1$), the decoder receives grounded future semantics for stable learning, while late in training ($\alpha\!\to\!0$), only the self-predicted forecast is supplied, thereby matching the inference regime.


\subsection{Planning Decoder}
\label{sec:method:dec}

Trajectory planning under closed-loop driving is inherently multi-modal: a deterministic regression head averages distinct intents into a single feasible-but-uninformative output. We therefore realise the planner as a conditional diffusion model so that multi-modal trajectory distributions emerge naturally, and route the future condition into every denoising step so that future semantics directly shape action generation rather than entering only as a post-hoc score.

\paragraph{Future-Conditioned Diffusion Transformer.}
The planning action is parameterised in the differential trajectory space $\mathbf{a}\!=\!(\Delta x,\Delta y,\sin\theta,\cos\theta)\!\in\!\mathbb{R}^{T\times 4}$. With a denoising diffusion probabilistic model (DDPM) forward process,
\begin{equation}
\mathbf{a}_{s}\;=\;\sqrt{\bar{\alpha}_{s}}\,\mathbf{a}_{0}\;+\;\sqrt{1-\bar{\alpha}_{s}}\,\boldsymbol{\epsilon},
\qquad \boldsymbol{\epsilon}\!\sim\!\mathcal{N}(\mathbf{0},\mathbf{I}),
\label{eq:ddpm_fwd}
\end{equation}
the noise predictor $\epsilon_{\theta}$ is a transformer in the diffusion-transformer (DiT) family that consumes a noisy action token, a timestep embedding $\mathbf{e}_{s}$, and two conditioning contexts, namely the scene context $\mathbf{C}_{\mathrm{scene}}\!=\![\mathbf{e}_{s}\,\Vert\,\mathbf{Z}_{t}]$ and the future context $\mathbf{Z}_{t+T}^{c}$:
\begin{equation}
\hat{\boldsymbol{\epsilon}}
\;=\;
\epsilon_{\theta}\!\left(\mathbf{a}_{s},\,s,\;\mathbf{C}_{\mathrm{scene}},\;\mathbf{Z}_{t+T}^{c}\right).
\label{eq:eps_pred}
\end{equation}
Each block first cross-attends to $\mathbf{C}_{\mathrm{scene}}$ and then to $\mathbf{Z}_{t+T}^{c}$, so that environmental geometry is consumed before the future-semantic correction is applied. To preserve a stable warm-start from a planner pretrained without future conditioning, the output projection of the future cross-attention is initialised to zero, making the freshly added module behave as an identity at the start of training and gradually learn to exploit $\mathbf{Z}_{t+T}^{c}$ as optimisation proceeds.

\subsection{Training}
\label{sec:method:train}
\paragraph{Trajectory Diffusion Objective.}
The primary supervision is the standard DDPM noise-prediction loss applied with the future-aware predictor in Eq.~\eqref{eq:eps_pred}:
\begin{equation}
\mathcal{L}_{\mathrm{plan}}
\;=\;
\mathbb{E}_{s,\boldsymbol{\epsilon},\mathbf{a}_{0}}
\left\|\boldsymbol{\epsilon} - \epsilon_{\theta}\!\left(\mathbf{a}_{s},s,\mathbf{C}_{\mathrm{scene}},\tilde{\mathbf{Z}}_{t+T}^{c}\right)\right\|_{2}^{2},
\label{eq:l_plan}
\end{equation}
where $\tilde{\mathbf{Z}}_{t+T}^{c}$ is the annealed future condition defined in Eq.~\eqref{eq:latentalign_anneal}. To prevent the BEV encoder from collapsing into representations that are useful only for trajectory regression, we retain a semantic auxiliary that decodes BEV tokens into a class-wise occupancy map and supervises it against the rasterised semantic ground truth $\mathbf{Y}^{\mathrm{bev}}$,
\begin{equation}
\mathcal{L}_{\mathrm{bev}}
\;=\;
\mathrm{CE}\!\left(\hat{\mathbf{Y}}^{\mathrm{bev}},\,\mathbf{Y}^{\mathrm{bev}}\right).
\label{eq:l_bev}
\end{equation}
The total objective is
\begin{equation}
\mathcal{L}\;=\;\lambda_{\mathrm{plan}}\,\mathcal{L}_{\mathrm{plan}}\;+\;\lambda_{\mathrm{bev}}\,\mathcal{L}_{\mathrm{bev}}.
\label{eq:l_total}
\end{equation}

\subsection{Inference}
\label{sec:method:infer}

At inference, the future observation $\mathbf{I}_{t+T}$ is unavailable. The Future Alignment Adapter is therefore bypassed, and the planning condition reduces directly to $\mathbf{Z}_{t+T}^{c} = \hat{\mathbf{Z}}_{t+T}$. This introduces a circular dependency: computing $\hat{\mathbf{Z}}_{t+T}$ via Eq.~\eqref{eq:wm_forward} requires a trajectory intent $\mathbf{E}_{\tau}$, which is itself the output being denoised. Progressive Foresight Guidance (PFG) resolves this dependency by supplying two phase-adaptive surrogate trajectory intents throughout the denoising process. The kinematic extrapolation $\boldsymbol{\tau}^{\mathrm{kin}}$ from Eq.~\eqref{eq:kin_extrap} is noise-free but geometrically coarse. A self-consistent estimate $\boldsymbol{\tau}^{\mathrm{tw}}$ is recovered from the current noisy sample via Tweedie's formula,
\begin{equation}
\hat{\mathbf{a}}_{0}^{(s)}
\;=\;
\frac{\mathbf{a}_{s}\;-\;\sqrt{1-\bar{\alpha}_{s}}\,\hat{\boldsymbol{\epsilon}}_{\varnothing}}{\sqrt{\bar{\alpha}_{s}}},
\qquad
\boldsymbol{\tau}^{\mathrm{tw}}
\;=\;
\mathrm{cumsum}\!\left(\hat{\mathbf{a}}_{0}^{(s)}\right),
\label{eq:tweedie}
\end{equation}
is reliable only once the noise level decreases. Let $r=s/(S\!-\!1)\in[0,1]$ index denoising progress. Three classifier-free guidance branches condition the Latent Dynamics Predictor on different trajectory intents:
\begin{equation}
\mathbf{Z}_{t+T}^{c,\varnothing}\!=\!f_{\psi}(\mathbf{Z}_{t};\,\mathbf{E}^{\varnothing}),\quad
\mathbf{Z}_{t+T}^{c,\mathrm{kin}}\!=\!f_{\psi}(\mathbf{Z}_{t};\,\phi_{\tau}(\boldsymbol{\tau}^{\mathrm{kin}})),\quad
\mathbf{Z}_{t+T}^{c,\mathrm{tw}}\!=\!f_{\psi}(\mathbf{Z}_{t};\,\phi_{\tau}(\boldsymbol{\tau}^{\mathrm{tw}})),
\label{eq:three_branches}
\end{equation}
yielding three noise predictions $\hat{\boldsymbol{\epsilon}}_{\varnothing},\hat{\boldsymbol{\epsilon}}_{\mathrm{kin}},\hat{\boldsymbol{\epsilon}}_{\mathrm{tw}}$ at each step. The guided noise estimate is a phase-dependent mixture
\begin{equation}
\hat{\boldsymbol{\epsilon}}
\;=\;
\hat{\boldsymbol{\epsilon}}_{\varnothing}\;+\;
w_{\mathrm{kin}}(r)\!\left(\hat{\boldsymbol{\epsilon}}_{\mathrm{kin}}\!-\!\hat{\boldsymbol{\epsilon}}_{\varnothing}\right)\;+\;
w_{\mathrm{tw}}(r)\!\left(\hat{\boldsymbol{\epsilon}}_{\mathrm{tw}}\!-\!\hat{\boldsymbol{\epsilon}}_{\varnothing}\right),
\label{eq:pfg_combine}
\end{equation}
with cosine schedules governed by a shared envelope template
\begin{equation}
w(r;\rho,\nu,w^{\max})
\;=\;
w^{\max}\cos\!\Bigl(\tfrac{\pi r}{2\rho}\Bigr)\mathbf{1}_{[r<\rho]}
\;+\;
\tfrac{w^{\max}}{2}\Bigl[1-\cos\!\Bigl(\tfrac{\pi(r-\nu)}{1-\nu}\Bigr)\Bigr]\mathbf{1}_{[r\ge\nu]},
\label{eq:pfg_schedule}
\end{equation}
where $w_{\mathrm{kin}}$ uses the decay form and $w_{\mathrm{tw}}$ uses the rise form. The two curves overlap on $(r_{\mathrm{start}}, r_{\mathrm{fade}})$, producing a smooth handover from the inertial prior to the self-consistent surrogate as the trajectory crystallises. Combined with the LatentAlign annealing of Sec.~\ref{sec:method:train}, the planning condition remains a future-aware latent at both training and inference, anchored to real future evidence in training and to phase-adaptive self-consistent surrogates at inference.

\section{Experiments}
\label{sec:experiments}
\subsection{Datasets and Evaluation Metrics}

We evaluate DriveFuture on the public \textit{NAVSIM} benchmark~\cite{navsim}, which is built on OpenScene~\cite{openscene2023} and nuPlan~\cite{caesar2022nuplan} logs for lightweight planning evaluation. We report results on \textit{NAVSIM-v1 navtest}~\cite{navsim}, \textit{NAVSIM-v2 navtest} ~\cite{navsimv2}, and \textit{NAVSIM-v2 navhard}~\cite{navsimv2}. \textit{navtest} measures general planning performance, while \textit{navhard} emphasizes safety-critical and long-tail scenarios.
Following the official protocol, we use PDMS for \textit{NAVSIM-v1} and EPDMS for \textit{NAVSIM-v2}. EPDMS extends PDMS with additional rule- and comfort-related metrics, including driving-direction compliance, traffic-light compliance, lane keeping, and extended comfort. For \textit{NAVSIM-v2 navhard}, we adopt the official two-stage evaluation, where Stage 2 re-evaluates the planner with synthesized future observations around the Stage-1 endpoint. All results are reported as percentages unless otherwise specified.


\begin{table}[t]
\centering
    \caption{Comparison with SOTA methods on the \textbf{NAVSIM-v2 navhard} split~\cite{navsim}.}
\renewcommand\arraystretch{0.9}
  \tabcolsep=2.3mm 
  \resizebox{\linewidth}{!}{
\begin{tabular}{l c c ccccc ccccc}
\toprule
\multicolumn{1}{l}{Method}&Backbone&Stage&  NC$\uparrow$& DAC$\uparrow$& DDC$\uparrow$& TL$\uparrow$& EP$\uparrow$& TTC$\uparrow$& LK$\uparrow$& HC$\uparrow$& EC$\uparrow$& EPDMS$\uparrow$ \\
        \midrule
    \rowcolor{cyan!15}\multicolumn{13}{c}{\textit{E2E-based Methods}} \\
        \multirow{2}{*}{TransFuser~\cite{chitta2022transfuser}} 
     & \multirow{2}{*}{ResNet-34} & Stage 1 & 96.2 & 79.5 & 99.1 & 99.5 & 84.1 & 95.1 & 94.2 & 97.5 & 79.1 & \multirow{2}{*}{23.1} \\
     &  & Stage 2 & 77.7 & 70.2 & 84.2 & 98.0 & 85.1 & 75.6 & 45.4 & 95.7 & 75.9 &  \\
    \arrayrulecolor{gray!25}\cmidrule[0.01em]{1-13}\arrayrulecolor{black}
        \multirow{2}{*}{DiffusionDrive~\cite{liao2025diffusiondrive}} 
     & \multirow{2}{*}{ResNet-34} & Stage 1 & 96.0 & 79.7 & 97.4 & 99.5 & 81.3 & 93.1 & 90.8 & 96.8 & 73.8 & \multirow{2}{*}{24.2} \\
     &  & Stage 2 & 82.1 & 72.2 & 88.5 & 98.7 & 85.1 & 78.8 & 49.2 & 89.3 & 71.2 &  \\
    \arrayrulecolor{gray!25}\cmidrule[0.01em]{1-13}\arrayrulecolor{black}
        \multirow{2}{*}{GuideFlow~\cite{liu2025guideflow}} & \multirow{2}{*}{ResNet-34} & Stage 1 & 96.6 & 80.5 & 96.3 & 99.3 & 82.3 & 94.9 & 91.5 & 97.7 & 67.8 & \multirow{2}{*}{27.1} \\
         &  & Stage 2 & 87.3 & 76.7 & 88.8 & 99.2 & 84.3 & 85.1 & 49.7 & 93.1 & 44.5 &  \\
        \arrayrulecolor{gray!25}\cmidrule[0.01em]{1-13}\arrayrulecolor{black}
        \multirow{2}{*}{Senna-E2E~\cite{jiang2024senna}} & \multirow{2}{*}{ResNet-50} & Stage 1 & 95.6 & 86.0 & 98.9 & 99.6 & 83.9 & 95.1 & 95.3 & 97.6 & 75.6 & \multirow{2}{*}{27.2} \\
         &  & Stage 2 & 78.6 & 74.8 & 84.8 & 98.2 & 88.2 & 75.7 & 46.9 & 96.0 & 65.8 &  \\
        \arrayrulecolor{gray!25}\cmidrule[0.01em]{1-13}\arrayrulecolor{black}
        \multirow{2}{*}{DriveSuprim~\cite{yao2025drivesuprim}} & \multirow{2}{*}{V2-99} & Stage 1 & 98.9 & 95.1 & 99.2 & 99.6 & 76.1 & 99.1 & 94.7 & 97.6 & 54.2 & \multirow{2}{*}{42.1} \\
         &  & Stage 2 & 87.9 & 88.8 & 89.6 & 98.8 & 80.3 & 86.0 & 53.5 & 97.1 & 56.1 &  \\
         \arrayrulecolor{gray!25}\cmidrule[0.01em]{1-13}\arrayrulecolor{black}
         \multirow{2}{*}{ZTRS~\cite{li2025ztrs}} & \multirow{2}{*}{V2-99} & Stage 1 & 98.9 & 97.6 & 100.0 & 100.0 & 66.7 & 98.9 & 96.2 & 96.7 & 44.0 & \multirow{2}{*}{48.1} \\
          &  & Stage 2 & 91.1 & 90.4 & 95.8 & 99.0 & 63.6 & 89.8 & 60.4 & 97.6 & 66.1 &  \\
         \arrayrulecolor{gray!25}\cmidrule[0.01em]{1-13}\arrayrulecolor{black}
         \multirow{2}{*}{GTRS-E~\cite{GTRS}} & \multirow{2}{*}{V2-99+EVA-ViT-L+ViT-L} & Stage 1 & 98.9 & 99.3 & 99.8 & 99.8 & 75.2 & 98.4 & 96.0 & 97.6 & 51.6 & \multirow{2}{*}{49.4} \\
          &  & Stage 2 & 92.3 & 93.3 & 94.6 & 99.2 & 73.1 & 91.2 & 53.9 & 96.7 & 56.8 &  \\
              \arrayrulecolor{gray!25}\cmidrule[0.01em]{1-13}\arrayrulecolor{black}
              \multirow{2}{*}{SimScale~\cite{tian2025simscale}} & \multirow{2}{*}{V2-99} & Stage 1 & 99.6 & 99.1 & 99.9 & 100.0 & 69.6 & 99.6 & 95.8 & 95.6 & 28.4 & \multirow{2}{*}{53.2} \\
               &  & Stage 2 & 94.5 & 94.2 & 95.8 & 99.2 & 75.8 & 92.8 & 60.1 & 96.1 & 43.2 &  \\
              \arrayrulecolor{gray!25}\cmidrule[0.01em]{1-13}\arrayrulecolor{black}
              \multirow{2}{*}{DrivoR~\cite{kirby2026drivor}} & \multirow{2}{*}{ViT-S} & Stage 1 & 99.1 & 98.2 & 99.3 & 99.8 & 75.4 & 98.7 & 94.9 & 97.6 & 70.2 & \multirow{2}{*}{54.6} \\
               &  & Stage 2 & 92.3 & 91.6 & 97.3 & 99.1 & 75.7 & 90.6 & 56.1 & 98.4 & 44.7 &  \\
        \midrule
         \rowcolor{cyan!15}\multicolumn{13}{c}{\textit{VLA-based Methods}} \\
         \multirow{2}{*}{SpanVLA~\cite{zhou2026spanvla}} & \multirow{2}{*}{Qwen2.5-VL-3B} & Stage 1 & 98.4 & 94.3 & 97.8 & 99.9 & 85.7 & 97.2 & 94.2 & 97.6 & 72.1 & \multirow{2}{*}{40.1} \\
          &  & Stage 2 & 86.9 & 84.3 & 87.1 & 98.2 & 85.5 & 82.7 & 62.3 & 96.8 & 67.4 &  \\
         \arrayrulecolor{gray!25}\cmidrule[0.01em]{1-13}\arrayrulecolor{black}
         \multirow{2}{*}{DiffVLA~\cite{jiang2025diffvla}} & \multirow{2}{*}{V2-99 + ViT-L/14} & Stage 1 & 95.7 & 99.2 & 100.0 & 100.0 & 85.9 & 96.4 & 97.1 & 95.0 & 84.2 & \multirow{2}{*}{45.0} \\
          &  & Stage 2 & 81.2 & 88.8 & 94.6 & 99.0 & 86.0 & 76.4 & 59.8 & 98.6 & 80.4 &  \\
         \arrayrulecolor{gray!25}\cmidrule[0.01em]{1-13}\arrayrulecolor{black}
        \midrule
        \rowcolor{cyan!15}\multicolumn{13}{c}{\textit{World-Model-based Methods}} \\
        \multirow{2}{*}{MindDrive~\cite{suna2025minddrive}} & \multirow{2}{*}{ResNet-34} & Stage 1 & 96.1 & 86.0 & 98.8 & 99.3 & 83.3 & 95.6 & 94.4 & 97.6 & 74.7 & \multirow{2}{*}{30.9} \\
         &  & Stage 2 & 82.6 & 79.1 & 86.4 & 98.0 & 85.3 & 79.4 & 49.2 & 96.5 & 71.0 &  \\
        \arrayrulecolor{gray!25}\cmidrule[0.01em]{1-13}\arrayrulecolor{black}
        \multirow{2}{*}{World4Drive~\cite{zheng2025world4drive}} & \multirow{2}{*}{ResNet-34} & Stage 1 & 97.3 & 89.1 & 97.6 & 99.7 & 60.5 & 96.8 & 87.7 & 93.1 & 60.0 & \multirow{2}{*}{34.9} \\
         &  & Stage 2 & 91.4 & 82.0 & 91.0 & 98.5 & 53.1 & 90.6 & 52.3 & 93.3 & 62.8 &  \\
        \arrayrulecolor{gray!25}\cmidrule[0.01em]{1-13}\arrayrulecolor{black}
        \rowcolor{red!15} & & Stage 1 & 99.8 & 99.8 & 100 & 99.6 & 85.7 & 99.8 & 98.7 & 97.6 & 66.2 &  \\
        \rowcolor{red!15} \multirow{-2}{*}{DriveFuture}& \multirow{-2}{*}{V2-99} & Stage 2 & 90.6 & 87.5 & 94.1 & 99.1 & 84.6 & 88.8 & 58.3 & 93.5 & 45.6 &  \multirow{-2}{*}{$\pmb{55.5}$}\\
\bottomrule
\end{tabular}}
\label{tab_navsimv2_navhard}
\end{table}

\begin{table}[t]
\centering
    \caption{Comparison with SOTA methods on the \textbf{NAVSIM-v2 navtest} split~\cite{navsim}. EPDMS$^{*}$ denotes results computed with the original NAVSIM-v2 evaluation code before the human-behavior filtering fix, while EPDMS denotes the corrected official implementation.}
\renewcommand\arraystretch{0.9}
  \tabcolsep=2.3mm 
  \resizebox{\linewidth}{!}{
\begin{tabular}{l c ccccc ccccc}
\toprule
\multicolumn{1}{l}{Method}&  NC$\uparrow$& DAC$\uparrow$& DDC$\uparrow$& TL$\uparrow$& EP$\uparrow$& TTC$\uparrow$& LK$\uparrow$& HC$\uparrow$& EC$\uparrow$& EPDMS$^{*}\uparrow$& EPDMS$\uparrow$ \\
        \midrule
        \rowcolor{cyan!15}\multicolumn{12}{c}{\textit{E2E-based Methods}} \\
        TransFuser~\cite{chitta2022transfuser} & 96.9 & 89.9 & 97.8 & 99.7 & 87.1 & 95.4 & 92.7 & 98.3 & 87.2 & 76.7 & -- \\
        DiffusionDrive~\cite{liao2025diffusiondrive} & 98.2 & 95.9 & 99.4 & 99.8 & 87.5 & 97.3 & 96.8 & 98.3 & 87.7 & -- & 84.5 \\
        Hydra-MDP++~\cite{li2025hydramdpp} & 97.2 & 97.5 & 99.4 & 99.6 & 83.1 & 96.5 & 94.4 & 98.2 & 70.9 & 81.4 & -- \\
        DriveSuprim~\cite{yao2025drivesuprim} & 97.5 & 96.5 & 99.4 & 99.6 & 88.4 & 96.6 & 95.5 & 98.3 & 77.0 & 83.1 & -- \\
        ARTEMIS~\cite{feng2025artemis} & 98.3 & 95.1 & 98.6 & 99.8 & 81.5 & 97.4 & 96.5 & 98.3 & 98.3 & 83.1 & -- \\
        DiffusionDriveV2~\cite{zou2025diffusiondrivev2} & 97.7 & 96.6 & 99.2 & 99.8 & 88.9 & 97.2 & 96.0 & 97.8 & 91.0 & 85.5 & 87.5 \\
        \midrule
        \rowcolor{cyan!15}\multicolumn{12}{c}{\textit{VLA-based Methods}} \\
        DriveWorld-VLA~\cite{liu2026driveworld} & 98.6 & 99.1 & 99.6 & 99.8 & 87.4 & 97.9 & 97.0 & 97.8 & 78.6 & -- & 86.8 \\
        DriveVLA-W0~\cite{li2025drivevla} & 98.5 & 99.1 & 98.0 & 99.7 & 86.4 & 98.1 & 93.2 & 97.9 & 58.9 & -- & 86.1 \\
        Recogdrive~\cite{li2025recogdrive} & 98.3 & 95.2 & 98.3 & 99.8 & 87.1 & 97.5 & 96.6 & 99.5 & 86.5 & -- & 83.6 \\
        \midrule
        \rowcolor{cyan!15}\multicolumn{12}{c}{\textit{World-Model-based Methods}} \\
        Latent-WAM~\cite{wang2026latentwam} & 98.1 & 97.3 & 99.6 & 99.8 & 87.7 & 97.3 & 97.6 & 98.1 & 87.3 & -- & 89.3 \\
        \rowcolor{red!15} DriveFuture & \textbf{98.8} & \textbf{99.1} & \textbf{99.6} & \textbf{99.9} & 86.6 & \textbf{98.4} & 96.4 & 98.3 & 74.8 & \textbf{86.4} & \textbf{89.9} \\
        \bottomrule
\end{tabular}}
\label{tab_navsimv2_navtest}
\end{table}

\subsection{Implementation Details}
\label{sec:exp_impl}
\begin{wraptable}{r}{0.5\textwidth}
\centering
\vspace{-0cm}
    \begin{minipage}{\linewidth}
    \centering
    \captionof{table}{Comparison with SOTA methods on the \textbf{NAVSIM-v1 navtest} split~\cite{navsim}.}
    \vspace{-0.0cm}
    \renewcommand\arraystretch{0.9}
    \tabcolsep=0.8mm 
    \resizebox{\linewidth}{!}{
    \begin{tabular}{l c c c c c c}
    \toprule
    \multicolumn{1}{l}{Method}&  NC$\uparrow$& DAC$\uparrow$& TTC$\uparrow$& Conf.$\uparrow$& EP$\uparrow$& PDMS$\uparrow$ \\
            \midrule
            Human &  100 & 100 & 100 & 99.9 & 87.5 & 94.8 \\
            Constant Velocity & 69.9 & 58.8 & 49.3 & 100 & 49.3 & 21.6 \\
            \midrule
            \rowcolor{cyan!15}\multicolumn{7}{c}{\textit{E2E-based Methods}} \\
            VADv2~\cite{chen2024vadv2} &  97.2 & 89.1 & 91.6 & 100 & 76.0 & 80.9 \\
            TransFuser~\cite{chitta2022transfuser} &  97.7 & 92.8 & 92.8 & 100 & 79.2 & 84.0 \\
            UniAD~\cite{uniad} &  97.8 & 91.9 & 92.9 & 100 & 78.8 & 83.4 \\
            PARA-Drive~\cite{weng2024para} &  97.9 & 92.4 & 93.0 & 99.8 & 79.3 & 84.0 \\
            DRAMA~\cite{yuan2024drama} &  98.0 & 93.1 & 94.8 & 100 & 80.1 & 85.5 \\
            GoalFlow~\cite{xing2025goalflow}  & 98.3 & 93.8 & 94.3 & 100 & 79.8 & 85.7 \\
            Hydra-MDP~\cite{li2024hydra} &  98.3 & 96.0 & 94.6 & 100 & 78.7 & 86.5 \\
            ARTEMIS~\cite{feng2025artemis} &  98.3 & 95.1 & 94.3 & 100 & 81.4 & 87.0 \\
            DiffusionDrive~\cite{liao2025diffusiondrive} & 98.2 & 96.2 & 94.7 & 100 & 82.2 & 88.1 \\
            DIVER~\cite{song2025diver} &  98.5 & 96.5 & 94.9 & 100 & 82.6 & 88.3 \\
            DriveSuprim~\cite{yao2025drivesuprim} &  97.8 & 97.3 & 93.6 & 100 & 86.7 & 89.9 \\
            GoalFlow~\cite{xing2025goalflow}  &  98.4 & 98.3 & 94.6 & 100 & 85.0 & 90.3 \\
            \midrule
            \rowcolor{cyan!15}\multicolumn{7}{c}{\textit{VLA-based Methods}} \\
           AutoVLA ~\cite{zhou2025autovla} & 98.4 & 95.6 & 98.0 & 99.9 & 81.9 & 89.1 \\
            Recogdrive~\cite{li2025recogdrive} &  98.2 & 97.8 & 95.2 & 99.8 & 83.5 & 89.6 \\
            DriveVLA-W0~\cite{li2025drivevla} &  98.7 & 99.1 & 95.3 & 99.3 & 83.3 & 90.2 \\
            DriveWorld-VLA~\cite{liu2026driveworld} &  99.1 & 98.2 & 96.1 & 100 & 85.9 & 91.3 \\
            \midrule
            \rowcolor{cyan!15}\multicolumn{7}{c}{\textit{World-Model-based Methods}} \\
            LAW~\cite{li2024enhancinglaw}  & 96.4 & 95.4 & 88.7 & 99.9 & 81.7 & 84.6 \\
            World4Drive~\cite{zheng2025world4drive}  & 97.4 & 94.3 & 92.8 & 100 & 79.9 & 85.1 \\
            WoTE~\cite{wote} &  98.5 & 96.8 & 94.9 & 99.9 & 81.9 & 88.3 \\
            WorldRFT~\cite{yang2026worldrft} & 97.8 & 96.8 & 94.0 & 100 & 81.7 & 87.8 \\
            DriveLaW~\cite{xia2025drivelaw} &  99.0 & 97.1 & 96.7 & 100 & 81.3 & 89.1 \\
            \rowcolor{red!15} DriveFuture & 98.8 & \textbf{99.1} & 95.4 & \textbf{100} & \textbf{84.2} & \textbf{90.7} \\
            \bottomrule
    \end{tabular}}
    \label{tab:navtest_v1_results}
\end{minipage}\hfill
\vspace{-1.cm}
\end{wraptable}
DriveFuture is built on a TransFuser-based~\cite{chitta2022transfuser_arxiv} planning pipeline, trained on the NavTrain split for 100 epochs with Adam at a learning rate of $10^{-4}$, and uses a GTRS-Dense scorer~\cite{GTRS} to select multi-modal trajectories. We use two temporal frames, front and rear camera streams, image resolution $2048\times512$, and a BEV token grid with $16\times64$ anchors. The planner predicts an $8$-step trajectory over a $4$ second horizon with a $0.5$ second interval. The diffusion planning head contains $5$ transformer layers and samples $100$ trajectory proposals by default.
Training is conducted on 8 NVIDIA 5090 GPUs with mixed bf16 precision, gradient clipping of $1.0$, and loss weights $\lambda_{\mathrm{plan}}=\lambda_{\mathrm{bev}}=10$. More implementation details are provided in the Supplementary Material.

\subsection{Main Results}
\label{sec:exp_main}



\noindent \textbf{NAVSIM-v2 navhard.}
As shown in Table~\ref{tab_navsimv2_navhard}, DriveFuture achieves the best overall result with \textbf{55.5} EPDMS, outperforming DrivoR~\cite{kirby2026drivor} (54.6 EPDMS), DiffVLA~\cite{jiang2025diffvla} (45.0 EPDMS), and the strongest prior world-model method World4Drive~\cite{zheng2025world4drive} (34.9 EPDMS). Notably, DriveFuture demonstrates consistent improvements across compliance and safety-related metrics, including 99.1 DAC and 95.4 TTC. These results indicate that conditioning the current latent state on future world states not only improves overall performance but also enhances safety-critical behavior under the challenging two-stage evaluation.

\noindent \textbf{NAVSIM-v2 navtest.} As shown in Table~\ref{tab_navsimv2_navtest}, DriveFuture achieves the best overall result of \textbf{89.9} corrected EPDMS. It surpasses the E2E-AD method DiffusionDriveV2~\cite{zou2025diffusiondrivev2} with 87.5, the VLA method DriveWorld-VLA~\cite{liu2026driveworld} with 86.8, and the world-model method Latent-WAM~\cite{wang2026latentwam} with 89.3. These gains indicate that DriveFuture improves planning robustness under the stricter NAVSIM-v2 protocol by explicitly conditioning current representations on future latent states.

\noindent \textbf{NAVSIM-v1 navtest.} As shown in Table~\ref{tab:navtest_v1_results}, DriveFuture achieves \textbf{90.7} PDMS. It surpasses GoalFlow~\cite{xing2025goalflow} with 90.3 and DriveLaW~\cite{xia2025drivelaw} with 89.1, while remaining competitive with the best VLA method DriveWorld-VLA~\cite{liu2026driveworld} at 91.3. This demonstrates that future-conditioned latent modeling better aligns world representations with planning requirements.

\begin{table}[H]
    \centering
    \caption{Unified ablation study on future supervision and PFG heuristic guidance on NAVSIM-v2 navhard~\cite{navsimv2}. A checkmark denotes the option is enabled in that configuration. FF denotes whether future frames are considered during training, Impl. denotes the proposed implicit future constraint, MSE denotes direct mean-squared-error supervision, KS denotes kinematics-based heuristic guidance, and GT denotes GT trajectory guidance. All ablations are conducted without GTRS-Dense scorer~\cite{GTRS}. }
    \label{tab:ablation_summary}
\renewcommand\arraystretch{0.9}
  \tabcolsep=4.3mm 
  \resizebox{\linewidth}{!}{
    \begin{tabular}{*{12}{c}}
        \toprule
        \multicolumn{3}{c}{Future Supervision} & \multicolumn{2}{c}{Heuristic Guidance} & \multirow{2}{*}{EPDMS} & \multirow{2}{*}{NC} & \multirow{2}{*}{DAC} & \multirow{2}{*}{TTC} & \multirow{2}{*}{LK} & \multirow{2}{*}{HC} & \multirow{2}{*}{EC} \\
        \cmidrule(lr){1-3}\cmidrule(lr){4-5}
        FF&Impl&MSE& KS & GT &  &  &  &  &  &  &  \\
        \midrule
         &  &  & \checkmark & \checkmark & 30.9 & 81.8 & 72.6 & 79.7 & 46.4 & 95.7 & 71.7 \\
        \checkmark &  & \checkmark & \checkmark & \checkmark & 32.1 & 81.7 & 77.0 & 78.4 & 49.5 & 96.7 & 69.9 \\
       \checkmark & \checkmark &  &  &  & 32.0 & 81.1 & 75.2 & 78.1 & 47.6 & 97.0 & 75.9 \\
        \rowcolor{red!15} \checkmark & \checkmark &  & \checkmark & \checkmark & \textbf{34.6} & \textbf{82.3} & \textbf{78.8} & \textbf{79.6} & \textbf{47.6} & \textbf{97.0} & \textbf{75.9} \\
        \bottomrule
    \end{tabular}}
\end{table}
\subsection{Ablation Study}
\label{sec:exp_ablation}



\textbf{Effect of Future Supervision.}
As shown in Table~\ref{tab:ablation_summary}, future-aware conditioning improves robustness, raising EPDMS from 30.9 to 34.6, with clear gains in drivable-area compliance and lane keeping. Direct future-latent MSE supervision is less effective, achieving 32.1 EPDMS, suggesting that future information is better used as an implicit planning condition than as a feature-level regression target.

\textbf{Effect of Heuristic Guidance .}
Table~\ref{tab:ablation_summary} evaluates the training sources used by PFG. When the kinematic branch and GT-guided training are removed, the model still produces reasonable proposals, but the final EPDMS drops from 34.6 to 32.0 and the largest deficits appear in Stage-2 DAC and TTC. This indicates that dual-source guidance is not only an inference heuristic: it also improves the quality of the learned future-conditioned latent during training and leads to more stable behavior in hard interactive rollouts.
\begin{wraptable}{r}{0.5\textwidth}
\vspace{-0.cm}
\centering
    \begin{minipage}{\linewidth}
    \centering
    \captionof{table}{Sensitivity analysis of three hyper-parameters on NAVSIM-v2 navhard~\cite{navsimv2}. We report the effect of future horizon $t_f$, query size $q_s$, and initial guidance parameter $e_o$ on final EPDMS. The best result in each group is highlighted in bold. Here $t_f$ denotes the $t_f$-th future frame, $q_s$ denotes the number of queries in the world model, and $e_o$ denotes the inflection point in the annealing schedule. All ablations are conducted without GTRS-Dense scorer~\cite{GTRS}. }
    \vspace{-0.cm}
    \label{tab:hyperparameter_sensitivity}
\renewcommand\arraystretch{0.95}
  \tabcolsep=3.9mm
  \resizebox{\linewidth}{!}{
    \begin{tabular}{c c c c c c}
        \toprule
        $t_f$ & EPDMS & $q_s$ & EPDMS & $e_o$ & EPDMS \\
        \midrule
        0.5 & 30.2 & 4 & 28.2 & 0.75 & 29.2 \\
        1.0 & 31.2 & \cellcolor{red!15}\textbf{16} & \cellcolor{red!15}\textbf{34.6} & \cellcolor{red!15}\textbf{0.83} & \cellcolor{red!15}\textbf{34.6} \\
        \cellcolor{red!15}\textbf{1.5} & \cellcolor{red!15}\textbf{34.6} & 64 & 33.7 & 0.95 & 28.9 \\
        \bottomrule
    \end{tabular}}
\end{minipage}\hfill
\vspace{-1.0cm}
\end{wraptable}


\textbf{Sensitivity of Hyper-parameters.}
Table~\ref{tab:hyperparameter_sensitivity} shows that moderate hyper-parameter settings are consistently preferred. The best EPDMS is achieved with $t_f=1.5$, $q_s=16$, and $e_o=0.83$ in their respective groups. Smaller or larger values reduce robustness, indicating that effective planning requires a balanced future horizon, query budget, and guidance strength.

\subsection{Qualitative Analysis}
\label{sec:exp_qualitative}
We analyze representative challenging cases from the NAVSIM-v2 \textit{navhard} split, as shown in Fig.~\ref{fig:qualitative_cases}. Compared with World4Drive\cite{zheng2025world4drive}, DriveFuture performs better in common failure cases, including collision, inefficient, and braking. This shows that conditioning current decision representations on future states is more effective than using future states merely as prediction targets.


\begin{figure}[t]
    \centering
    \includegraphics[width=1\linewidth]{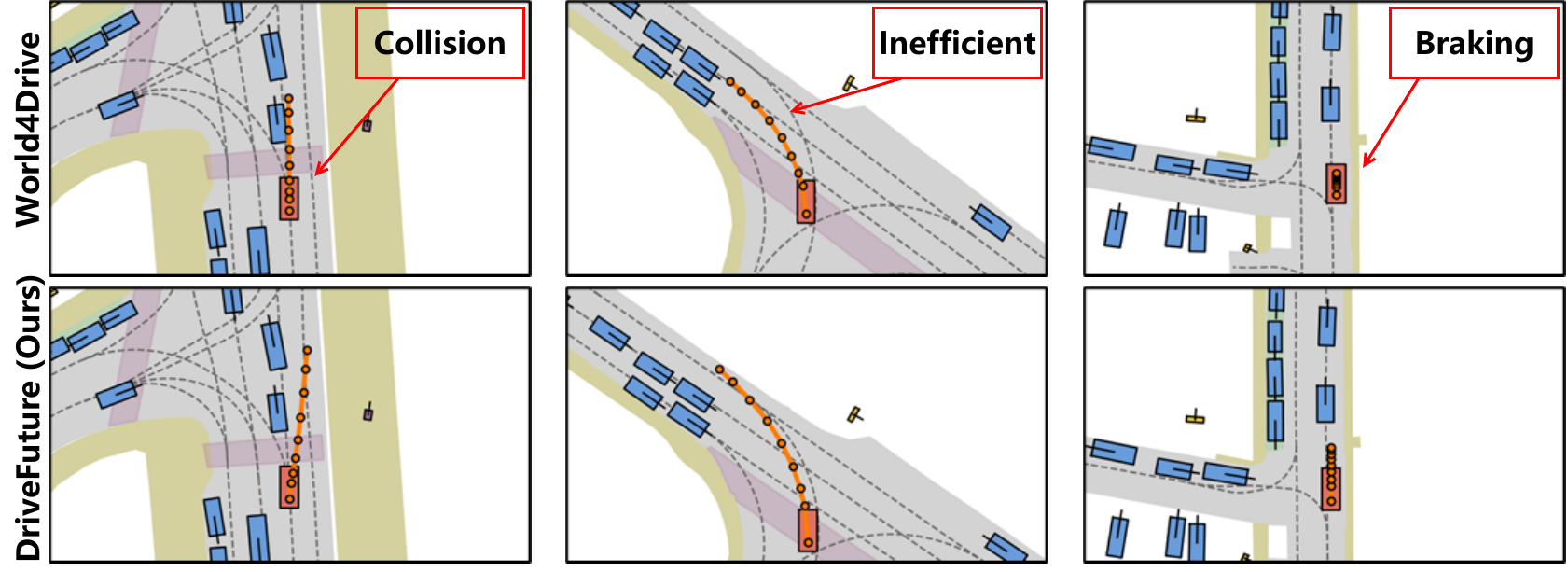}
    \caption{Visual comparison between latent world model World4Drive~\cite{zheng2025world4drive} and our DriveFuture across multiple driving scenarios from NAVSIM-v2 navhard~\cite{navsim}.}
    \label{fig:qualitative_cases}
\end{figure}


\section{Conclusion}
In this work, we propose \textbf{DriveFuture}, a future-aware latent world modeling framework for autonomous driving. Unlike existing methods that primarily simulate future states, DriveFuture uses future world states as explicit conditions to shape current latent representations for planning. It consists of a Latent Dynamics Predictor for compact future latent prediction, and a Future Alignment Adapter for extracting planning-relevant information from GT future observations. DriveFuture achieves SOTA performance on public NAVSIM-v1/v2 benchmarks, validating the effectiveness of future-conditioned latent world modeling for autonomous driving.

\textbf{Limitation and Future Work.}
DriveFuture depends on predicted future latents during inference, so inaccurate future prediction may affect planning in uncertain scenarios. It also mainly focuses on near-future conditioning. In the future, we will explore uncertainty-aware and longer-horizon future modeling to improve robustness and generalization.



\bibliographystyle{unsrtnat}
\bibliography{main}
\newpage

\appendix
\section{Experiment Details}
\label{app:experiment_details}

\subsection{Datasets and Benchmarks}
\label{app:datasets_benchmarks}

\textbf{Datasets.}
We train and evaluate DriveFuture on the nuPlan (OpenScene) data used by the public \textit{NAVSIM} benchmark~\cite{navsim}. The nuPlan dataset~\cite{caesar2022nuplan} provides large-scale real-world autonomous driving logs with multi-camera observations, ego states, HD-map information, object annotations, and human driving trajectories. OpenScene~\cite{openscene2023} serves as a compact redistribution of nuPlan, and NAVSIM further organizes these logs into a lightweight planning-oriented benchmark. In our experiments, NAVSIM provides real-world driving scenes, historical ego states, map context, and future trajectories for training and evaluating end-to-end planners. 

\textbf{Benchmarks.}
Following the official NAVSIM protocol, we report results on three settings: \textit{NAVSIM-v1 navtest}, \textit{NAVSIM-v2 navtest}, and \textit{NAVSIM-v2 navhard}. The \textit{navtest} split evaluates general planning ability on standard real-world driving scenes, while \textit{navhard} focuses on more challenging and safety-critical long-tail scenarios. \textit{NAVSIM-v1} uses the Predictive Driver Model Score (PDMS) to assess key aspects of driving behavior, including safety, feasibility, comfort, and progress. \textit{NAVSIM-v2} adopts the Extended Predictive Driver Model Score (EPDMS), which further incorporates rule- and comfort-related metrics such as driving-direction compliance, traffic-light compliance, lane keeping, and extended comfort. For \textit{NAVSIM-v2 navhard}, we follow the official two-stage protocol: Stage 1 evaluates the planner on the original observation, and Stage 2 re-evaluates it with synthesized future observations around the Stage-1 endpoint. All reported scores are percentages, and higher values indicate better planning quality.

\subsection{Evaluation Metrics}
\label{app:evaluation_metrics}

\textbf{PDM and PDMS}. Following NAVSIM~\cite{navsim}, the Predictive Driver Model (PDM) refers to the rule-based planner used to generate and score trajectory proposals. Given an observation $o_t$ and a set of candidate trajectories $\mathcal{T}=\{\tau_i\}_{i=1}^{N}$, PDM selects the trajectory with the highest PDM score:
\begin{equation}
\tau^{\star}=\arg\max_{\tau_i\in\mathcal{T}} \mathrm{Score}_{\mathrm{PDM}}(\tau_i),
\label{eq:appendix_pdm_selection}
\end{equation}
where $\mathrm{Score}_{\mathrm{PDM}}$ evaluates each candidate by simulating it and aggregating safety, feasibility, progress, and comfort terms. NAVSIM adopts this PDM-style scoring function as the benchmark metric, namely the Predictive Driver Model Score (PDMS).

For \textit{NAVSIM-v1}, the primary metric is PDMS. The evaluation first unrolls the predicted trajectory in a non-reactive simulator and computes normalized subscores in $[0,1]$. These subscores are then divided into hard penalties and soft objectives. The hard penalties are no-at-fault collision (NC) and drivable-area compliance (DAC), while the soft objectives are ego progress (EP), time-to-collision (TTC), and comfort (C). The final score is computed as
\begin{equation}
\mathrm{PDMS}
=
\underbrace{\prod_{m\in\{\mathrm{NC},\mathrm{DAC}\}} s_m}_{\text{hard penalties}}
\cdot
\underbrace{
\frac{\sum_{w\in\{\mathrm{EP},\mathrm{TTC},\mathrm{C}\}} \alpha_w s_w}
{\sum_{w\in\{\mathrm{EP},\mathrm{TTC},\mathrm{C}\}} \alpha_w}
}_{\text{weighted soft score}},
\label{eq:appendix_pdms}
\end{equation}
where $s_m$ denotes the normalized subscore and NAVSIM uses $\alpha_{\mathrm{EP}}=5$, $\alpha_{\mathrm{TTC}}=5$, and $\alpha_{\mathrm{C}}=2$. The multiplicative penalty term makes safety-critical violations dominate the final score: if the ego trajectory causes an at-fault collision, $s_{\mathrm{NC}}$ becomes zero; if the trajectory leaves the drivable area, $s_{\mathrm{DAC}}$ becomes zero. Collisions with static objects are assigned a softer penalty in NAVSIM, while non-at-fault collisions under the non-reactive setting are ignored.

The soft terms measure driving quality when the trajectory is admissible. The ego-progress subscore $s_{\mathrm{EP}}$ is computed as the ratio between the ego progress along the route centerline and a safe upper-bound progress estimated by the privileged PDM-Closed planner, clipped to $[0,1]$. The time-to-collision subscore $s_{\mathrm{TTC}}$ is initialized to one and is set to zero if, at any simulation step within the 4-second horizon, the projected ego motion violates the predefined TTC safety threshold with respect to surrounding vehicles. The comfort subscore $s_{\mathrm{C}}$ evaluates whether the trajectory satisfies acceleration and jerk thresholds. Therefore, PDMS rewards trajectories that simultaneously remain collision-free, stay on the road, make sufficient route progress, preserve safety margins, and maintain smooth motion.

For \textit{NAVSIM-v2}, the benchmark extends PDMS to the Extended Predictive Driver Model Score (EPDMS) by adding more fine-grained rule-compliance and comfort terms. In addition to NC, DAC, EP, and TTC, EPDMS includes driving-direction compliance (DDC), traffic-light compliance (TLC), lane keeping (LK), history comfort (HC), and extended comfort (EC). Following the NAVSIM-v2 protocol~\cite{navsimv2}, the single-stage extended score can be written as
\begin{equation}
\mathrm{EPDMS}
=
\underbrace{
\prod_{m\in\mathcal{M}_{\mathrm{pen}}}
f_m(\tau_{\mathrm{agent}},\tau_{\mathrm{human}})
}_{\text{penalty terms}}
\cdot
\underbrace{
\frac{
\sum_{m\in\mathcal{M}_{\mathrm{avg}}}\beta_m
f_m(\tau_{\mathrm{agent}},\tau_{\mathrm{human}})
}{
\sum_{m\in\mathcal{M}_{\mathrm{avg}}}\beta_m
}
}_{\text{weighted average terms}},
\label{eq:appendix_epdms}
\end{equation}
where $\mathcal{M}_{\mathrm{pen}}=\{\mathrm{NC},\mathrm{DAC},\mathrm{DDC},\mathrm{TLC}\}$ and $\mathcal{M}_{\mathrm{avg}}=\{\mathrm{EP},\mathrm{TTC},\mathrm{LK},\mathrm{HC},\mathrm{EC}\}$. The default weights are $\beta_{\mathrm{EP}}=5$, $\beta_{\mathrm{TTC}}=5$, and $\beta_{\mathrm{LK}}=\beta_{\mathrm{HC}}=\beta_{\mathrm{EC}}=2$. Here, $f_m(\tau_{\mathrm{agent}},\tau_{\mathrm{human}})$ denotes the filtered subscore for metric $m$. This filtering mechanism ignores a rule violation if the same violation is also committed by the human trajectory in the corresponding scene, reducing false penalties caused by annotation noise or contextually necessary maneuvers. DDC evaluates whether the ego vehicle follows the legal driving direction, TLC checks obedience to traffic lights, LK evaluates lane-keeping behavior, and HC/EC measure trajectory smoothness under the extended NAVSIM-v2 protocol.

We distinguish between EPDMS$^{*}$ and EPDMS when reporting NAVSIM-v2 results. EPDMS$^{*}$ denotes scores computed with the earlier NAVSIM-v2 evaluation implementation before the human-behavior filtering fix was adopted in the official leaderboard. It preserves the same extended metric set as Eq.~\eqref{eq:appendix_epdms}, but may penalize the agent for violations that are also present in the human reference behavior. EPDMS denotes the corrected official implementation, where the human-filtered subscores $f_m(\cdot)$ are used consistently. Therefore, EPDMS is the primary metric for final comparison, while EPDMS$^{*}$ is reported only for compatibility with earlier results computed using the legacy code.

For \textit{NAVSIM-v2 navhard}, evaluation follows the pseudo-simulation protocol with two stages. Stage 1 scores the planner from the original real observation, producing $s_1$. Stage 2 evaluates the planner on a set of pre-generated synthetic observations around plausible future ego states, producing scores $\{s_2^{(i)}\}_{i=1}^{K}$. These Stage-2 scores are aggregated by a Gaussian-weighted average according to the distance between each synthetic start point $x_i$ and the Stage-1 endpoint $\hat{x}$:
\begin{equation}
s_2=\sum_{i=1}^{K}\hat{w}_i s_2^{(i)},
\quad
\hat{w}_i=\frac{w_i}{\sum_{j=1}^{K}w_j},
\quad
w_i=\exp\left(-\frac{\lVert x_i-\hat{x}\rVert_2^2}{2\sigma^2}\right).
\label{eq:appendix_stage2_weighting}
\end{equation}
The final navhard score multiplies the original-observation score and the aggregated synthetic-observation score:
\begin{equation}
\mathrm{EPDMS}_{\mathrm{navhard}} = s_1\cdot s_2.
\label{eq:appendix_navhard_epdms}
\end{equation}
This two-stage aggregation evaluates both immediate planning quality and robustness to future observation shifts, making EPDMS on navhard stricter than single-stage NAVSIM-v1 PDMS or NAVSIM-v2 navtest evaluation.

\section{More Details on DriveFuture}
\label{app:more_details_drivefuture}

This section provides additional technical details omitted from the main paper due to space constraints. We first give a more explicit formulation of DriveFuture as a future-aware latent world model for autonomous driving planning, then expand the derivation of the Progressive Foresight Guidance (PFG) sampler, and finally discuss why the proposed training--inference design is better suited to planning than conventional future-prediction objectives. DriveFuture builds on the NAVSIM planning stack~\cite{navsim,navsimv2} and is motivated by recent work on latent world models~\cite{li2024enhancinglaw,zheng2025world4drive,yang2026worldrft,wote,liu2026driveworld,li2025drivevla,xia2025drivelaw,wang2026latentwam} and trajectory diffusion planners~\cite{liao2025diffusiondrive,song2025diver}.

\subsection{More Details of Notation}
\label{app:notation}

Let $\mathbf{I}_t$ denote the multi-view camera observations and $\mathbf{s}_t$ the ego-status at scene time $t$. DriveFuture first maps $(\mathbf{I}_t, \mathbf{s}_t)$ to a compact current latent representation
\begin{equation}
  \mathbf{Z}_t = \phi_{\mathrm{enc}}(\mathbf{I}_t, \mathbf{s}_t) \in \mathbb{R}^{N\times d},
  \quad N=65,\quad d=256,
  \label{eq:app_encoder}
\end{equation}
where $64$ tokens correspond to spatial BEV anchors and the final token encodes the ego-status. The planner predicts a trajectory
\begin{equation}
  \tau = \{(x_k,y_k,\theta_k)\}_{k=1}^{T}, \quad T=8,
  \label{eq:app_traj_def}
\end{equation}
covering a $4$ second horizon at $0.5$ second intervals. A trajectory tokenizer $\phi_\tau$ maps an absolute trajectory to $T$ tokens using normalised finite differences and sine/cosine heading embeddings:
\begin{align}
  \Delta x_k &= x_k - x_{k-1}, \quad \Delta y_k = y_k - y_{k-1}, \quad x_0=y_0=0, \\
  \bar{\Delta x}_k &= \frac{\Delta x_k-\mu_x}{\sigma_x}, \quad
  \bar{\Delta y}_k = \frac{\Delta y_k-\mu_y}{\sigma_y}, \\
  \phi_\tau(\boldsymbol{\tau})_k &= \operatorname{LN}\!\left(W_{\tau}[\bar{\Delta x}_k,\bar{\Delta y}_k,\sin\theta_k,\cos\theta_k]^{\top}+\mathbf{p}_k\right).
  \label{eq:app_tokenizer}
\end{align}
The world model then predicts a compact foresight latent:
\begin{equation}
  \hat{\mathbf{Z}}_{t+T}=f_{\psi}\bigl(\mathbf{Z}_t;\mathbf{E}_\tau\bigr)\in\mathbb{R}^{K\times d},
  \quad K=16.
  \label{eq:app_world_model_interface}
\end{equation}
The planning decoder is a diffusion transformer that predicts the noise of a noised trajectory sample while cross-attending both the current latent and the foresight latent:
\begin{equation}
  \epsilon_\theta=\epsilon_\theta\bigl(\mathbf{a}_s,s,\mathbf{Z}_t,\hat{\mathbf{Z}}_{t+T}\bigr).
  \label{eq:app_denoiser_interface}
\end{equation}

Throughout the appendix, $(\mathbf{I}_t,\mathbf{s}_t)$ denotes the current multi-camera observation and ego-status, and $\mathbf{Z}_t\in\mathbb{R}^{65\times256}$ denotes the corresponding current scene latent with $64$ BEV tokens plus one status token. The planned trajectory $\tau$ contains $T=8$ poses over a $4$ second horizon, and $\mathbf{E}_\tau=\phi_\tau(\boldsymbol{\tau})$ denotes its tokenized representation. The latent dynamics predictor $f_{\psi}$ maps the current scene and trajectory intent into a compact foresight latent $\hat{\mathbf{Z}}_{t+T}$ with $K=16$ tokens. During training, $\mathbf{Z}_{t+T}$ is the stop-gradient future BEV token bank, and $\tilde{\mathbf{Z}}_{t+T}$ is the Future Alignment Adapter output obtained by querying this bank with the predicted foresight latent. At inference, $\mathbf{E}^{\varnothing}$ denotes the null trajectory token for unconditional CFG, $\boldsymbol{\tau}^{\mathrm{kin}}$ denotes the constant-acceleration rollout, $\mathbf{a}_s$ is the noisy trajectory representation at diffusion step $s$, and $\epsilon_\theta$ is the denoiser prediction.

\subsection{More Details on Future-Aware Latent World Modeling}
\label{app:future_conditioned_formulation}

Most latent world models for autonomous driving optimize a future-prediction objective of the form
\begin{equation}
  \min_{\psi} \; \mathbb{E}\left[D\bigl(f_{\psi}(\mathbf{Z}_t,a_t),\mathbf{Z}^{\star}_{t+T}\bigr)\right],
  \label{eq:app_future_prediction}
\end{equation}
where $D$ is a feature-space distance or reconstruction loss. Such objectives are useful for modeling environment dynamics, but they do not necessarily enforce that the future latent contains information that is useful for choosing a trajectory. DriveFuture changes the role of future information: instead of only asking whether the future latent can be predicted, it asks whether the future latent improves denoising of the planned trajectory.

This leads to a planning-oriented objective:
\begin{equation}
  \min_{\theta,\psi,\omega}\; \mathbb{E}_{\tau,s,\epsilon}\left[
  \left\|\epsilon-\epsilon_\theta\bigl(\mathbf{a}_s,s,\phi_{\mathrm{enc}}(\mathbf{I}_t,\mathbf{s}_t),f_{\psi}(\mathbf{Z}_t;\mathbf{E}_\tau)\bigr)\right\|_2^2
  \right] + \lambda_{\mathrm{bev}}\mathcal{L}_{\mathrm{BEV}}.
  \label{eq:app_planning_oriented_obj}
\end{equation}
The important distinction is that gradients from the trajectory denoising loss directly shape the world model. Thus, the predicted future latent is not required to reconstruct every visual detail of the future scene. It only needs to preserve the future information that the planner can exploit, such as lane occupancy, route feasibility, collision-relevant motion, and interactions that affect the ego trajectory.

This treatment differs from several common uses of future information in end-to-end driving. Trajectory-generation methods often represent the future through sampled trajectories, goals, or action distributions, so future reasoning mainly appears at the output level after the current representation has already been formed. Trajectory-scoring methods generate multiple proposals and then rank them, which can improve selection but does not necessarily change how the proposals are represented. Pixel-, video-, or occupancy-space world models predict dense future observations, but this can be expensive and may allocate capacity to planning-irrelevant appearance details. Latent world models avoid some of this cost, yet their future states are often optimized as prediction targets or auxiliary signals. DriveFuture instead uses compact planning-oriented future tokens as a direct conditioning pathway for the trajectory diffusion denoiser, and this pathway can further complement downstream trajectory scoring.

\subsection{More Details on Latent Dynamics Predictor}
\label{app:world_model_architecture}

The latent dynamics predictor uses a Transformer decoder with learnable future queries. Let
\begin{equation}
  \mathbf{C}=\left[\mathbf{Z}_t\;\|\;\mathbf{E}_\tau\right]\in\mathbb{R}^{(N+T)\times d}
  \label{eq:app_context_concat}
\end{equation}
be the concatenated context. Given learnable future queries $\mathbf{Q}^{+,0}\in\mathbb{R}^{K\times d}$, each layer performs
\begin{align}
  \tilde{\mathbf{Q}}^{+,\ell} &= \operatorname{SelfAttn}\left(\operatorname{LN}(\mathbf{Q}^{+,\ell-1})\right)+\mathbf{Q}^{+,\ell-1}, \\
  \bar{\mathbf{Q}}^{+,\ell} &= \operatorname{CrossAttn}\left(Q=\operatorname{LN}(\tilde{\mathbf{Q}}^{+,\ell}),K=V=\mathbf{C}\right)+\tilde{\mathbf{Q}}^{+,\ell}, \\
  \mathbf{Q}^{+,\ell} &= \operatorname{FFN}\left(\operatorname{LN}(\bar{\mathbf{Q}}^{+,\ell})\right)+\bar{\mathbf{Q}}^{+,\ell}.
  \label{eq:app_wm_decoder_layer}
\end{align}
After $L_{\mathrm{wm}}=4$ decoder layers, $\hat{\mathbf{Z}}_{t+T}=\mathbf{Q}^{+,L_{\mathrm{wm}}}$. The compact $K$-token ($K=16$) output is a deliberate bottleneck: it prevents the world model from copying dense future appearance and encourages it to encode future information that has high utility for planning.

\subsection{More Details on Conditioning Source Randomisation}
\label{app:three_mode_conditioning}

DriveFuture trains the world model under three conditioning modes:
\begin{equation}
  \mathbf{E}_\tau \sim
  \begin{cases}
    \phi_\tau(\boldsymbol{\tau}^{\mathrm{gt}}) & \text{w.p.}\;p_{\mathrm{gt}},\\[2pt]
    \phi_\tau(\boldsymbol{\tau}^{\mathrm{kin}}) & \text{w.p.}\;p_{\mathrm{kin}},\\[2pt]
    \mathbf{E}^\varnothing & \text{w.p.}\;p_{\varnothing}.
  \end{cases}
  \label{eq:app_three_modes}
\end{equation}
In our default configuration, $(p_{\mathrm{gt}},p_{\mathrm{kin}},p_{\varnothing})=(0.4,0.4,0.2)$. The three modes have complementary roles. The expert trajectory provides a high-quality future-intent signal during training, the kinematic rollout exposes the model to a deployable coarse intent, and the null token creates the unconditional branch required by classifier-free guidance. The null branch is especially important: without an unconditional baseline, guidance would be an uncalibrated difference between two conditional predictions rather than a proper correction direction.

The kinematic rollout is computed using a constant-acceleration ego model. Given current ego velocity $v=(v_x,v_y)$ and acceleration $a=(a_x,a_y)$, the $k$-th future pose at interval $\Delta t$ is
\begin{align}
  t_k &= k\Delta t, \\
  x_k^{\mathrm{kin}} &= v_x t_k + \frac{1}{2}a_x t_k^2, \\
  y_k^{\mathrm{kin}} &= v_y t_k + \frac{1}{2}a_y t_k^2, \\
  \theta_k^{\mathrm{kin}} &= \operatorname{atan2}(v_y+a_y t_k, v_x+a_x t_k).
  \label{eq:app_kinematic_rollout}
\end{align}
This rollout is not intended to be an accurate planner. Its purpose is to provide a stable, physically plausible intent direction when the diffusion sample is still too noisy to be trusted.

\subsection{More Details on Future Alignment Adapter}
\label{app:future_alignment_details}

During training, DriveFuture can access the future observation $\mathbf{I}_{t+T}$. It is encoded with the same perception backbone under a stop-gradient operation:
\begin{equation}
  \mathbf{Z}_{t+T}=\operatorname{sg}\left(\phi_{\mathrm{enc}}(\mathbf{I}_{t+T})\right)\in\mathbb{R}^{64\times d}.
  \label{eq:app_future_gt_encode}
\end{equation}
Rather than directly using all $64$ future tokens as denoiser condition, DriveFuture lets the predicted foresight latent query the future token bank:
\begin{align}
  \mathbf{A} &= \operatorname{softmax}\left(\frac{\operatorname{LN}(\hat{\mathbf{Z}}_{t+T})W_Q(\mathbf{Z}_{t+T}W_K)^{\top}}{\sqrt{d_h}}\right),\\
  \tilde{\mathbf{Z}}_{t+T} &= \mathbf{A}\,\mathbf{Z}_{t+T}W_V.
  \label{eq:app_future_interact_expanded}
\end{align}
This makes the future oracle trajectory-aware: the query side is the world model's compact prediction, while the value side is the real future BEV. The module has no residual connection from $\hat{\mathbf{Z}}_{t+T}$, so the oracle condition is a pure selection of ground-truth future evidence.

The curriculum mixture is
\begin{equation}
  \tilde{\mathbf{Z}}_{t+T}^{c}=\alpha(e)\mathbf{Z}_{t+T}^{c}+(1-\alpha(e))\hat{\mathbf{Z}}_{t+T}.
  \label{eq:app_curriculum_mix}
\end{equation}
In the code implementation used for the current experiments, the schedule is equivalently written as
\begin{equation}
  \alpha(e)=1-\frac{1}{1+\exp\left[-\beta(e-e_0)\right]},
  \quad e_0=\rho_E E,
  \label{eq:app_alpha_code}
\end{equation}
where $E$ is the configured maximum epoch. At early epochs $e\ll e_0$, $\alpha(e)\approx 1$, and the denoiser learns with a strong future oracle. At late epochs $e\gg e_0$, $\alpha(e)\approx 0$, forcing the world model prediction to carry the conditioning information used at inference.

\subsection{More Details on Progressive Foresight Guidance}
\label{app:dspcfg}

At inference, the expert trajectory $\boldsymbol{\tau}^{\mathrm{gt}}$ and future observation $\mathbf{I}_{t+T}$ are unavailable. The Future Alignment Adapter is therefore bypassed and $\mathbf{Z}_{t+T}^{c}=\hat{\mathbf{Z}}_{t+T}$. A circular dependency then arises:
\begin{equation}
  \hat{\mathbf{Z}}_{t+T}=f_{\psi}\bigl(\mathbf{Z}_t;\mathbf{E}_\tau\bigr),
  \qquad
  \boldsymbol{\tau}=\operatorname{Decode}\bigl(\mathbf{Z}_t,\hat{\mathbf{Z}}_{t+T}\bigr).
  \label{eq:app_circularity}
\end{equation}
The trajectory intent $\mathbf{E}_\tau$ is required to compute $\hat{\mathbf{Z}}_{t+T}$, yet $\boldsymbol{\tau}$ is itself the output of the denoising process conditioned on $\hat{\mathbf{Z}}_{t+T}$. PFG breaks this circularity by supplying two deployable surrogate sources: an external coarse source $\boldsymbol{\tau}^{\mathrm{kin}}$ and an internal self-estimated source $\boldsymbol{\tau}^{\mathrm{tw}}$.

For a diffusion step $s$, define the unconditional, kinematic, and trajectory-conditioned future latents as
\begin{align}
  \mathbf{Z}_{t+T}^{c,\varnothing} &= f_{\psi}(\mathbf{Z}_t,\mathbf{E}^\varnothing), \\
  \mathbf{Z}_{t+T}^{c,\mathrm{kin}} &= f_{\psi}(\mathbf{Z}_t,\phi_\tau(\boldsymbol{\tau}^{\mathrm{kin}})), \\
  \mathbf{Z}_{t+T}^{c,\mathrm{tw}} &= f_{\psi}(\mathbf{Z}_t,\phi_\tau(\boldsymbol{\tau}^{\mathrm{tw}})).
  \label{eq:app_three_infer_latents}
\end{align}
The guided denoising direction is
\begin{equation}
  \hat{\boldsymbol{\epsilon}} = \hat{\boldsymbol{\epsilon}}_\varnothing
  + w_{\mathrm{kin}}(r)\left(\hat{\boldsymbol{\epsilon}}_{\mathrm{kin}}-\hat{\boldsymbol{\epsilon}}_\varnothing\right)
  + w_{\mathrm{tw}}(r)\left(\hat{\boldsymbol{\epsilon}}_{\mathrm{tw}}-\hat{\boldsymbol{\epsilon}}_\varnothing\right),
  \label{eq:app_ds_guidance}
\end{equation}
where $r\in[0,1]$ is denoising progress. This can be interpreted as a first-order guidance composition around the unconditional score estimate. The two correction terms induce two different directions in trajectory space: a stable low-frequency motion prior from kinematics and a high-precision self-consistency prior from the denoised sample.

The Tweedie estimate used by the second source is
\begin{equation}
  \hat{\mathbf{a}}_0^{(s)}=\frac{\mathbf{a}_s-\sqrt{1-\bar{\alpha}_s}\,\hat{\boldsymbol{\epsilon}}_\varnothing}{\sqrt{\bar{\alpha}_s}}.
  \label{eq:app_tweedie_expanded}
\end{equation}
If $\hat{\boldsymbol{\epsilon}}_\varnothing=\epsilon+\delta$, then the error of the Tweedie estimate is
\begin{equation}
  \hat{\mathbf{a}}_0^{(s)}-\mathbf{a}_0 = -\sqrt{\frac{1-\bar{\alpha}_s}{\bar{\alpha}_s}}\,\delta.
  \label{eq:app_tweedie_error}
\end{equation}
This explains why the trajectory source should be suppressed at high noise: when $\bar{\alpha}_s$ is small, the error amplification factor is large. Conversely, at low noise, $\bar{\alpha}_s\rightarrow1$ and the estimate becomes reliable. The progressive schedule implements this observation.

\begin{align}
  w_{\mathrm{kin}}(r) &=
  \begin{cases}
  w^{\max}_{\mathrm{kin}}\cos\!\bigl(\tfrac{\pi r}{2\rho}\bigr), & r<\rho,\\
  0, & r\ge \rho,
  \end{cases}
  \label{eq:app_wkin}\\
  w_{\mathrm{tw}}(r) &=
  \begin{cases}
  0, & r\le \nu,\\
  \tfrac{w^{\max}_{\mathrm{tw}}}{2}\Bigl[1-\cos\!\Bigl(\pi\tfrac{r-\nu}{1-\nu}\Bigr)\Bigr], & r>\nu.
  \end{cases}
  \label{eq:app_wtraj}
\end{align}
The default values are $w^{\max}_{\mathrm{kin}}=1.5$, $w^{\max}_{\mathrm{tw}}=2.5$, $\rho=0.7$, and $\nu=0.3$.

The schedule has an intuitive phase structure. At high noise ($r<\nu$), $\mathbf{a}_s$ is close to Gaussian and Tweedie estimates are unstable, so PFG mainly relies on the kinematic source to stabilize coarse direction and avoid implausible early drift. At middle noise ($\nu<r<\rho$), the denoised estimate becomes partially meaningful while kinematics still regularizes the solution, producing a smooth handover between the coarse prior and the self-consistent plan. At low noise ($r>\rho$), $\hat{\mathbf{a}}_0^{(s)}$ becomes reliable and reflects the denoiser's current mode, so the Tweedie trajectory source dominates final refinement through trajectory-aware future latents.

\subsection{More Details on Training Objective}
\label{app:full_objective}

The complete training objective used in DriveFuture is
\begin{equation}
  \mathcal{L}=\lambda_{\mathrm{plan}}\mathcal{L}_{\mathrm{plan}}+
  \lambda_{\mathrm{bev}}\mathcal{L}_{\mathrm{BEV}},
  \quad \lambda_{\mathrm{plan}}=\lambda_{\mathrm{bev}}=10.
  \label{eq:app_total_loss}
\end{equation}
The diffusion planning loss is
\begin{equation}
  \mathcal{L}_{\mathrm{plan}}=\mathbb{E}_{s,\epsilon}\left\|\epsilon-
  \epsilon_\theta\left(\sqrt{\bar{\alpha}_s}\mathbf{a}_0+
  \sqrt{1-\bar{\alpha}_s}\epsilon,
  s,\mathbf{Z}_t,\tilde{\mathbf{Z}}_{t+T}^{c}\right)\right\|_2^2.
  \label{eq:app_dp_loss}
\end{equation}
The BEV semantic loss is a seven-class cross-entropy loss over the upsampled BEV prediction:
\begin{equation}
  \mathcal{L}_{\mathrm{BEV}}=-\frac{1}{|\Omega|}\sum_{u\in\Omega}\sum_{c=1}^{7}y_{u,c}\log p_{u,c}.
  \label{eq:app_bev_loss}
\end{equation}
The BEV objective anchors the latent scene representation to metric semantics and stabilizes trajectory diffusion training. It also makes the latent space more compatible with future-frame alignment because both current and future BEV tokens are produced by the same encoder family.

\section{Implementation Details for DriveFuture}
\label{app:implementation_details_drivefuture}

This section describes the implementation used for the submitted DriveFuture models. The details are reconstructed from the training and evaluation scripts, the Hydra configuration, and the model code. We include them to facilitate reproducibility and to clarify several engineering choices that materially affect performance on NAVSIM-v1 and NAVSIM-v2.

\subsection{More Details on Model Configuration}

DriveFuture uses two temporal frames and two stitched panorama streams as input. The front panorama concatenates left-front, front, and right-front camera views, while the rear panorama concatenates left-rear, rear, and right-rear views; both panoramas are resized to $2048\times512$. The image encoder follows a VoVNet / V2-99-style backbone initialized from detection pretraining. Its image features are queried by learnable BEV anchors arranged on a $16\times64$ grid, which are downscaled into $64$ spatial BEV tokens. A single ego-status token is appended to these BEV tokens, yielding the current latent $\mathbf{Z}_t\in\mathbb{R}^{65\times256}$. The world model uses $16$ future queries, $4$ Transformer decoder layers, and an FFN dimension of $2048$. The trajectory head predicts $8$ poses over a $4$ second horizon at $0.5$ second intervals. The diffusion planning head contains $5$ Transformer layers and samples $100$ proposals at inference. For PFG, the default guidance strengths are $(w^{\max}_{\mathrm{kin}},w^{\max}_{\mathrm{tw}})=(1.5,2.5)$ with phase parameters $(\rho,\nu)=(0.7,0.3)$. The world model and diffusion head share the same latent dimension $d=256$, which avoids projection mismatch between current scene tokens, future tokens, and trajectory tokens.

\subsection{More Details on Training Configuration}

DriveFuture is trained on the NAVSIM navtrain split with Adam, a learning rate of $1\times10^{-4}$, batch size $16$ per step, and gradient accumulation over $5$ steps. Development runs use $100$--$130$ epochs, bf16 mixed precision, gradient clipping at $1.0$, epoch-level checkpointing, and online Weights \& Biases logging. The train loader uses roughly $7$--$9$ workers, the validation loader uses $2$ workers. For efficiency, training is performed in cache-only mode with precomputed NAVSIM features: the data loader reads cached features and targets rather than reconstructing the scene loader online, and cache-on-miss behavior is disabled in the final training run to avoid accidental recomputation. The same training step also uses the default conditioning-source dropout $(p_{\mathrm{gt}},p_{\mathrm{kin}},p_{\varnothing})=(0.4,0.4,0.2)$. Algorithm~\ref{alg:app_training_step} summarizes how these settings are instantiated in each DriveFuture training step.

\begin{algorithm}[t]
\caption{DriveFuture training step}
\label{alg:app_training_step}
\small
\begin{algorithmic}[1]
\Require Batch of features $F$, targets $Y$, current epoch $e$
\State $\mathbf{Z}_t,\mathbf{B}_t \leftarrow \mathcal{E}_{\theta}(F)$ \Comment{Current BEV and semantic features}
\State $\tau_{\mathrm{kin}} \leftarrow \textsc{KinematicRollout}(F.s_t)$
\State $\hat{\mathbf{Z}}^{+}_{t+T}\leftarrow\mathcal{W}_{\psi}(\mathbf{Z}_t,\phi(Y.\tau^{\star}),\phi(\tau_{\mathrm{kin}}),\mathbf{c}^{\varnothing})$ \Comment{Three-mode dropout internally selects the condition}
\If{future frame exists}
  \State $\mathbf{Z}^{+,\star}_{t+T}\leftarrow \operatorname{sg}(\mathcal{E}_{\theta}(Y.o_{t+T}))$
  \State $c_{\mathrm{interact}}\leftarrow\operatorname{CrossAttn}(\hat{\mathbf{Z}}^{+}_{t+T},\mathbf{Z}^{+,\star}_{t+T})$
  \State $\tilde{\mathbf{Z}}^{+}\leftarrow\alpha(e)c_{\mathrm{interact}}+(1-\alpha(e))\hat{\mathbf{Z}}^{+}_{t+T}$
\Else
  \State $\tilde{\mathbf{Z}}^{+}\leftarrow\hat{\mathbf{Z}}^{+}_{t+T}$
\EndIf
\State Sample diffusion step $s$ and noise $\epsilon$
\State $\mathbf{a}_s\leftarrow\sqrt{\bar{\alpha}_s}\phi(Y.\tau^{\star})+\sqrt{1-\bar{\alpha}_s}\epsilon$
\State $\hat{\epsilon}\leftarrow\epsilon_{\omega}(\mathbf{a}_s,s,\mathbf{Z}_t,\tilde{\mathbf{Z}}^{+})$
\State $\mathcal{L}\leftarrow 10\|\epsilon-\hat{\epsilon}\|_2^2+10\mathcal{L}_{\mathrm{BEV}}(\mathbf{B}_t,Y_{\mathrm{BEV}})$
\State \Return $\mathcal{L}$
\end{algorithmic}
\end{algorithm}

\subsection{More Details on Inference}

Algorithm~\ref{alg:app_inference} summarizes the inference pipeline used by DriveFuture, including CFG guidance and the optional scorer-based final selection. At inference, DriveFuture generates $100$ trajectory proposals per scene. Each proposal starts from independent Gaussian noise. The current scene latent $\mathbf{Z}_t$, unconditional future latent $\mathbf{Z}^{+}_{\varnothing}$, and kinematic future latent $\mathbf{Z}^{+}_{\mathrm{kin}}$ are computed once per scene and repeated across proposals. The Tweedie future latent $\mathbf{Z}^{+}_{\mathrm{tw}}(s)$ is recomputed only on denoising steps where $w_{\mathrm{tw}}(r)>0$.

\begin{algorithm}[t]
\caption{DriveFuture inference with PFG and optional trajectory scoring}
\label{alg:app_inference}
\small
\begin{algorithmic}[1]
\Require Observation $o_t$, ego-status $s_t$, proposal count $K=100$
\State $\mathbf{Z}_t\leftarrow\mathcal{E}_{\theta}(o_t,s_t)$
\State $\tau_{\mathrm{kin}}\leftarrow\textsc{KinematicRollout}(s_t)$
\State $\mathbf{Z}^{+}_{\varnothing}\leftarrow\mathcal{W}_{\psi}(\mathbf{Z}_t,\mathbf{c}^{\varnothing})$
\State $\mathbf{Z}^{+}_{\mathrm{kin}}\leftarrow\mathcal{W}_{\psi}(\mathbf{Z}_t,\phi(\tau_{\mathrm{kin}}))$
\For{$k=1$ to $K$}
  \State Sample $\mathbf{a}_{S}^{(k)}\sim\mathcal{N}(0,I)$
  \For{$s=S$ to $1$}
    \State Compute $w_{\mathrm{kin}}(r)$ and $w_{\mathrm{tw}}(r)$
    \State Predict $\epsilon_{\varnothing}$ under $\mathbf{Z}^{+}_{\varnothing}$
    \State Add kinematic correction if $w_{\mathrm{kin}}>0$
    \State Estimate $\hat{\mathbf{a}}_0$ by Tweedie and add trajectory correction if $w_{\mathrm{tw}}>0$
    \State Update $\mathbf{a}_{s-1}^{(k)}$ with the diffusion scheduler
  \EndFor
  \State $\tau^{(k)}\leftarrow\textsc{CumSum}(\mathbf{a}_0^{(k)})$
\EndFor
\If{using GTRS-Dense scorer}
  \State $k^{\star}\leftarrow\arg\max_k \operatorname{Score}_{\mathrm{GTRS}}(\tau^{(k)},o_t)$
  \State \Return $\tau^{(k^{\star})}$ and proposals $\{\tau^{(k)}\}_{k=1}^{K}$
\Else
  \State \Return default proposal and proposals $\{\tau^{(k)}\}_{k=1}^{K}$
\EndIf
\end{algorithmic}
\end{algorithm}

For NAVSIM-v2 navhard leaderboard submission, we use DriveFuture proposals with a GTRS-Dense scorer~\cite{GTRS}. This follows the recent observation that high-quality proposal scoring is crucial for NAVSIM performance~\cite{GTRS,yao2025drivesuprim,li2025ztrs,sun2026sparsedrivev2}. DriveFuture improves the proposal distribution, while the scorer selects the best candidate under a stronger planning metric proxy. The main ablations toggle the runnable implementation through several switches: \texttt{use\_wm} enables the latent dynamics predictor, \texttt{use\_wm\_to\_dit} feeds the foresight latent into the diffusion transformer, \texttt{use\_interact} enables the training-time future interaction adapter, and \texttt{force\_alpha\_one} tests the effect of keeping the oracle future condition throughout training. At inference, \texttt{use\_dspcfg} enables PFG, \texttt{use\_kinematic\_extrap} enables the constant-acceleration guidance source, and \texttt{p\_gt}, \texttt{p\_kin}, and \texttt{p\_null} control the three-mode CFG dropout with default values $0.4/0.4/0.2$. These switches connect the runnable configuration directly to the ablation study.

\section{More Results}
\label{app:more_results}

This section expands the experimental discussion in the main paper. We provide additional breakdowns, relative improvements, ablation-oriented analysis, and interpretation of the main metrics. The three settings stress different properties. NAVSIM-v1 navtest emphasizes the original PDM-style safety and progress metrics. NAVSIM-v2 navtest adds rule compliance and comfort terms. NAVSIM-v2 navhard is the strictest setting because the final score multiplies Stage-1 quality by robustness to Stage-2 synthesized future observations. DriveFuture is designed for precisely this type of evaluation: the world model explicitly uses future-conditioned latent representations, and PFG improves proposal generation under uncertain future evolution.

\subsection{More Details on Navhard}

Table~\ref{tab:app_navhard_relative} quantifies how much DriveFuture + Score improves over representative NAVSIM-v2 navhard baselines, while Table~\ref{tab:app_drivefuture_navhard_stagewise} breaks the same result down into Stage-1 and Stage-2 metrics for DriveFuture with and without scoring.

\begin{wraptable}{r}{0.5\textwidth}
\centering
\vspace{-0.3cm}
    \begin{minipage}{\linewidth}
    \centering
    \captionof{table}{Relative comparison on NAVSIM-v2 navhard using the main-paper results. Improvements are computed against the listed method's EPDMS.}
    \label{tab:app_navhard_relative}
    \renewcommand\arraystretch{0.9}
      \tabcolsep=2.3mm 
      \resizebox{\linewidth}{!}{%
    \begin{tabular}{l c c c}
    \toprule
    \textbf{Method} & \textbf{EPDMS} & \textbf{Gain} & \textbf{Rel.} \\
    \midrule
    TransFuser~\cite{chitta2022transfuser} & 23.1 & +32.4 & +140.3\% \\
    DiffusionDrive~\cite{liao2025diffusiondrive} & 24.2 & +31.3 & +129.3\% \\
    GuideFlow~\cite{liu2025guideflow} & 27.1 & +28.4 & +104.8\% \\
    MindDrive~\cite{suna2025minddrive} & 30.9 & +24.6 & +79.6\% \\
    World4Drive~\cite{zheng2025world4drive} & 34.9 & +20.6 & +59.0\% \\
    GTRS-E~\cite{GTRS} & 49.4 & +6.1 & +12.3\% \\
    SimScale~\cite{tian2025simscale} & 53.2 & +2.3 & +4.3\% \\
    DrivoR~\cite{kirby2026drivor} & 54.6 & +0.9 & +1.6\% \\
    \bottomrule
    \end{tabular}}
\end{minipage}\hfill
\vspace{-0.4cm}
\end{wraptable}

The relative gains in Table~\ref{tab:app_navhard_relative} are largest over models without strong two-stage robustness or dense proposal scoring. More importantly, DriveFuture maintains an advantage even over strong recent methods that already use trajectory scoring or larger backbones. This suggests that improving the proposal distribution remains complementary to stronger downstream ranking.

Table~\ref{tab:app_drivefuture_navhard_stagewise} shows that the benefit of scoring is not confined to the easier first stage. The addition of scoring increases Stage-1 safety metrics almost to saturation: NC and DAC both reach $99.8$, while DDC reaches $100.0$. On Stage 2, where future observations are perturbed, the scorer also substantially improves NC, DAC, DDC, TTC, and LK. The lower EC value for the scored model indicates a known trade-off in NAVSIM-style planning: selecting safer and more rule-compliant proposals can reduce the extended comfort metric when the chosen trajectory is more conservative or involves stronger braking. Since EPDMS uses multiplicative penalties for safety-critical metrics, the safety and compliance improvements dominate the final score.

\begin{table}[H]
\centering
\caption{DriveFuture stage-wise NAVSIM-v2 navhard metrics from the main paper. Stage 1 evaluates the original observation; Stage 2 evaluates synthesized follow-up observations.}
\label{tab:app_drivefuture_navhard_stagewise}
\renewcommand\arraystretch{0.9}
  \tabcolsep=4.3mm 
\resizebox{\linewidth}{!}{%
\begin{tabular}{l c c c c c c c c c c}
\toprule
\textbf{Model} & \textbf{Stage} & NC & DAC & DDC & TL & EP & TTC & LK & HC & EC \\
\midrule
DriveFuture & Stage 1 & 96.3 & 87.8 & 98.2 & 99.6 & 83.1 & 96.9 & 94.9 & 97.6 & 76.9 \\
DriveFuture & Stage 2 & 82.3 & 78.8 & 88.1 & 98.4 & 83.6 & 79.6 & 47.6 & 97.0 & 75.9 \\
DriveFuture + Score & Stage 1 & 99.8 & 99.8 & 100.0 & 99.6 & 85.7 & 99.8 & 98.7 & 97.6 & 66.2 \\
DriveFuture + Score & Stage 2 & 90.6 & 87.5 & 94.1 & 99.1 & 84.6 & 88.8 & 58.3 & 93.5 & 45.6 \\
\bottomrule
\end{tabular}%
}
\end{table}

\subsection{More Details on NAVSIM-v2 Navtest}

\label{app:qualitative_results}

\begin{figure}[t]
    \centering
    \includegraphics[width=1\linewidth]{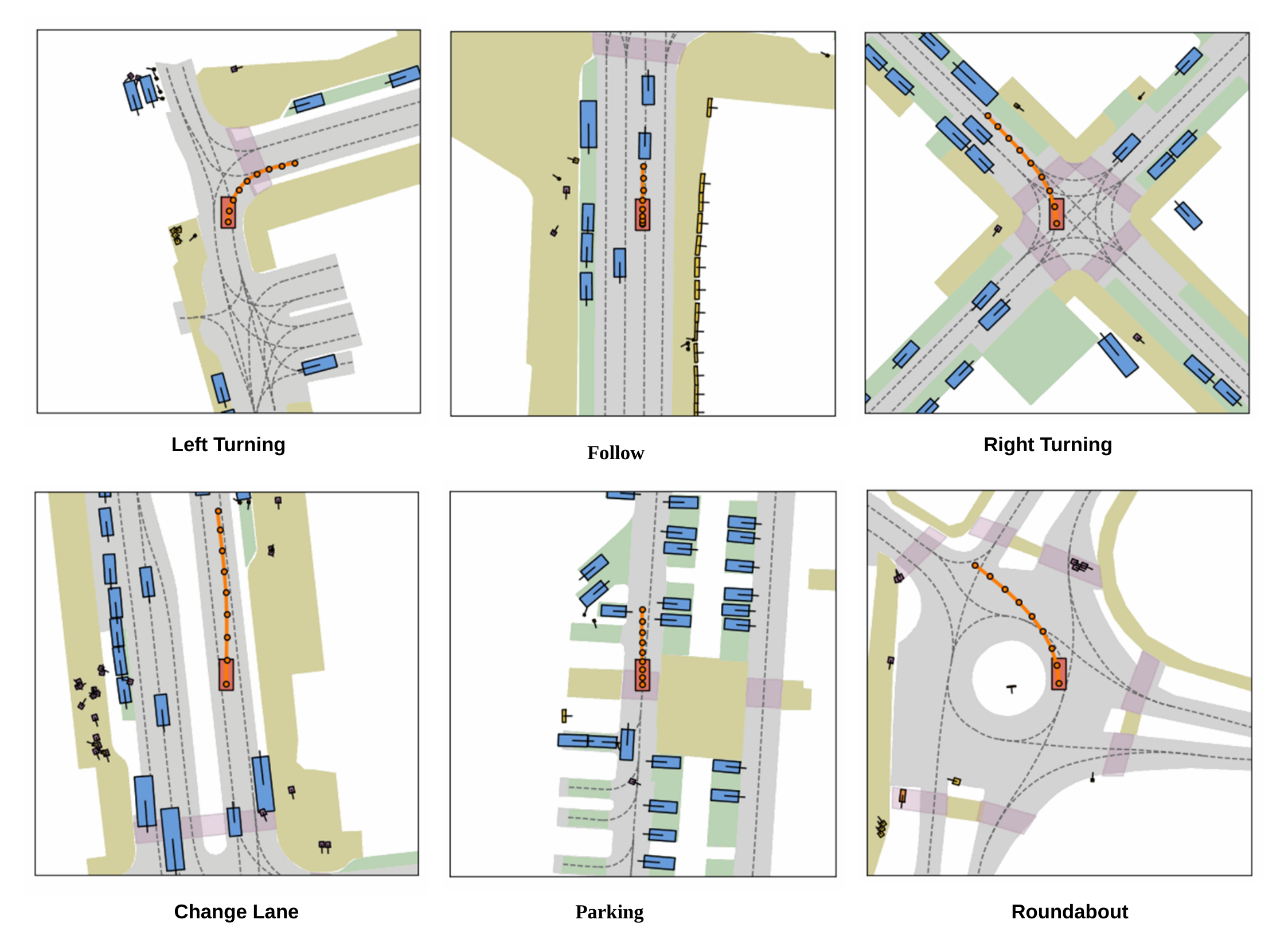}
    \caption{ DriveFuture across multiple driving scenarios from NAVSIM-v2 navhard.~\cite{navsim}.}
    \label{fig:qualitative_cases}
    \vspace{-0.2cm}
\end{figure}

Although we do not repeat the full NAVSIM-v2 navtest leaderboard table in this appendix, the main-paper results show that DriveFuture is particularly strong on NC, DAC, DDC, TL, and TTC. This metric pattern is consistent with the method design: future-conditioned latent planning improves the model's ability to avoid near-future unsafe states, while PFG stabilizes proposal generation before final scoring. The corrected EPDMS is higher than EPDMS$^{*}$ because the official human-behavior filtering fix avoids penalizing the agent for violations that also occur in the human reference behavior.

\subsection{More Details on NAVSIM-v1 Navtest}

The NAVSIM-v1 score shows a similar pattern: DriveFuture is strongest on feasibility and safety-related terms. Ego progress is not maximized relative to some aggressive methods, but the final PDMS remains strong because hard penalties such as NC and DAC have multiplicative influence. This is desirable for autonomous driving: a planner that makes slightly less progress while preserving collision-free and drivable-area behavior is often preferred by PDM-style metrics. This metric-level pattern is consistent with the role of each DriveFuture component. The world model $\mathcal{W}$ adds a compact foresight latent and prevents the planner from degenerating into current-scene-only diffusion. The Future Alignment Adapter supplies high-fidelity future BEV evidence during training, and the latent-alignment curriculum transfers this oracle signal into the predicted future latent while reducing train--test mismatch. Three-mode CFG dropout calibrates the GT, kinematic, and null branches within one world model. During inference, the kinematic source stabilizes high-noise denoising, the Tweedie source refines low-noise samples with self-consistent trajectory intent, and the GTRS-Dense scorer selects the strongest proposal from the generated set. Removing these components is expected to weaken future awareness, branch calibration, early denoising stability, final proposal refinement, or candidate selection, respectively.

Fig.~\ref{fig:qualitative_cases} provides visual evidence that is consistent with the quantitative gains reported in the main paper. From left to right, the top row corresponds to curved turning, straight lane following with nearby traffic, and turning through a complex intersection, while the bottom row shows dense-traffic lane following, traversal through a narrow road segment constrained by parked vehicles, and roundabout navigation. Despite substantial variation in road topology, traffic density, and interaction complexity, DriveFuture consistently produces trajectories that remain well aligned with the underlying road structure and establish reasonable turning or forward-progress trends at an early stage. The predicted trajectories exhibit smoother curvature transitions and fewer artifacts such as visible jitter, over-steering, or ineffective lateral drift. In the more interactive scenes, the model also adapts earlier to conflict-zone geometry and implicit safety boundaries. Taken together, these qualitative patterns support the main quantitative results and suggest that the future-conditioned latent improves not only benchmark scores, but also the coherence, safety, and topology consistency of the resulting planning behavior.


\end{document}